\documentclass[lettersize,journal]{IEEEtran}
\usepackage{amsmath,amsfonts}
\usepackage{algorithmic}
\usepackage{algorithm}
\usepackage{array}
\usepackage[caption=false,font=normalsize,labelfont=sf,textfont=sf]{subfig}
\usepackage{textcomp}
\usepackage{stfloats}
\usepackage{url}
\usepackage{verbatim}
\usepackage{graphicx}
\usepackage{cite}
\usepackage{multirow}
\usepackage{tabularx}
\usepackage{booktabs}
\usepackage{dsfont}
\usepackage{tablefootnote}
\usepackage{hyperref}
\begin{document}

\title{Integrated Conflict Management for UAM with Strategic Demand Capacity Balancing and Learning-based Tactical Deconfliction}

% \title{Learning-based Tactical Deconfliction with Demand-Capacity Balancing Coordination}
\author{Shulu Chen, Antony Evans, Marc Brittain and Peng Wei
\thanks{S. Chen is with the Department of Electrical and Computer Engineering, George Washington University, Washington, DC 20052 \href{mailto:shulu@gwu.edu}{shulu@gwu.edu}}
\thanks{A. Evans is the Director of System Design for Airbus UTM at Acubed, an Airbus innovation center, Sunnyvale, CA 94086 \href{mailto:tony.evans@airbus-sv.com}{tony.evans@airbus-sv.com}}
\thanks{M. Brittain is a member of the Technical Staff at MIT Lincoln Laboratory, Lexington, MA 02421 \href{mailto:marc.brittain@ll.mit.edu}{marc.brittain@ll.mit.edu}}
\thanks{P. Wei is with the Department of Mechanical and Aerospace Engineering, George Washington University, Washington, DC 20052 \href{mailto:pwei@gwu.edu}{pwei@gwu.edu}}}



\maketitle

\begin{abstract}
Urban air mobility (UAM) has the potential to revolutionize our daily transportation, offering rapid and efficient deliveries of passengers and cargo between dedicated locations within and around the urban environment. 
Before the commercialization and adoption of this emerging transportation mode, however, aviation safety must be guaranteed, i.e., all the aircraft have to be safely separated by strategic and tactical deconfliction.
Reinforcement learning has demonstrated effectiveness in the tactical deconfliction of en route commercial air traffic in simulation. However, its performance is found to be dependent on the traffic density. 
In this project, we propose a novel framework that combines demand capacity balancing (DCB) for strategic conflict management and reinforcement learning for tactical separation. 
By using DCB to precondition traffic to proper density levels, we show that reinforcement learning can achieve much better performance for tactical safety separation. Our results also indicate that this DCB preconditioning can allow target levels of safety to be met that are otherwise impossible. 
In addition, combining strategic DCB with reinforcement learning for tactical separation can meet these safety levels while achieving greater operational efficiency than alternative solutions. 
\end{abstract}


\section{Introduction}
\subsection{Motivation}
According to projections, the number of air vehicles operating in urban areas will experience a significant increase in the next two decades \cite{balakrishnan2018blueprint,jenkins2017forecast,hamilton_2018}.
One major part of this forecasted traffic surge is from electric vertical take-off and landing (eVTOL) cargo and passenger air taxis in Urban Air Mobility (UAM) operations.
The current Air Traffic Control (ATC) system is heavily human-based, which is not expected to support the emerging high-density urban air traffic operations \cite{national2014autonomy}. 
Automation tools and autonomous agents to manage the urban airspace and UAM traffic are required.
Autonomous ATC was proposed in 2005 with the introduction of the NASA Advanced Airspace Concept (AAC) \cite{erzberger2004transforming}. 
This rule-based autonomous ATC tool was further developed and validated over the following 10 years to augment human ATC, increase traffic capacity and enhance operation safety\cite{erzberger2010algorithm, erzberger2014design}.  

Specifically for UAM, the US Federal Aviation Administration (FAA) and National Aeronautics and Space Administration (NASA) proposed concepts for Unmanned Aircraft System (UAS) Traffic Management (UTM) in recent years \cite{kopardekar2016unmanned, thipphavong2018urban,faa2020uam, faa2023uam}. 
From these proposals, one of the most challenging requirements for an autonomous ATC system is to mitigate conflicts in high-density traffic flows. 
This can be achieved through a combination of strategic conflict management, which is used to resolve predicted conflicts prior to departure by adding a ground delay or rescheduling another flight route, and tactical deconfliction, which focuses on real-time decision making for airborne aircraft separation through maneuver advisories like speed or heading changes. 

Various autonomous conflict management systems have been developed, but one persistent challenge in the integration of such systems is to ensure the advisories are coordinated to achieve the desired safety level.
If this is not the case, the strategic and tactical deconfliction methods may affect each other's results and introduce new risks.  
To address this issue, we propose an integrated conflict management framework (ICMF) that combines both strategic conflict management and tactical deconfliction. 
By implementing this comprehensive autonomous system in air traffic management (ATM) for UAM, we seek to guarantee safety levels within target values, while also optimizing traffic efficiency.

\subsection{Related Work}

Strategic conflict management involves strategic decisions like ground delays made by air traffic managers to balance traffic demand with airspace capacity at bottlenecks, e.g. airport runways, merging points, and air route intersections. 
For traditional ATM, such an approach has been designed effectively and has shown measurable improvements for airlines in the National Airspace System (NAS). 
For example, Traffic Management Initiatives (TMI) such as the Ground Delay Program (GDP), Airspace Flow Program (AFP), and the Collaborative Trajectory Options Program (CTOP) are tools used by air traffic flow managers to balance demand with capacity in congested regions \cite{zhu2019decision}. 
These programs have resulted in reduced delays and cancellations for airlines operating in the NAS, while also improving safety levels by reducing the number of aircraft in the airspace and preventing potential conflicts. 
However, strategic conflict management for UAM is still a challenge because of the high-density traffic and high-population areas over which that traffic operates \cite{chen2022estimating}. 
Therefore, further research is required to study and analyze the effectiveness of strategic conflict management in the UAM setting, specifically with the integration of tactical deconfliction technologies.

The field of aircraft separation assurance has seen the introduction of many advanced methods, as highlighted by recent studies \cite{razzaghi2022survey}. 
One such approach involves using Markov Decision Processes (MDP) to formulate the separation assurance problem by incorporating a probabilistic model that can handle uncertainties encountered during flight \cite{account}. 
Offline MDP-based methods are useful for strategic deconfliction, while online MDP-based methods are more suitable for tactical deconfliction \cite{ong2017markov, bertram2020distributed, taye2022reachability}. 
However, offline methods can become intractable if uncertainty occurs en route since the policy is designed ahead of time, and it is challenging for online methods to solve the problem efficiently \cite{scalableDRL}. 
To address these challenges, researchers have turned to deep reinforcement learning (DRL) for separation assurance problems~\cite{brittain2019autonomous, D2MAV, scalableDRL, D2MAV-A, guo2021safety}. 
For instance, the deep distributed multi-agent variable (D2MAV-A) framework incorporates an attention network and employs a modified Proximal Policy Optimization (PPO) algorithm to solve complex sequential decision-making problems with a variable number of agents \cite{D2MAV-A}. 
Nevertheless, a key concern with DRL is its generalization ability - if the density of traffic flow exceeds the training environment, the DRL agent may provide erroneous advisories and lead to an aircraft conflict, or even a near mid-air collision. 
Thus, preconditioning air traffic to proper density levels using strategic conflict management is essential for DRL to ensure safe separation.

\subsection{Contributions and Structure}

The major contributions of this paper are summarized as follows:
\begin{enumerate}
\item{\bf An integrated conflict management framework for UAM.}  
This new framework is a coordination between strategic conflict management and tactical deconfliction.
Through our analysis, we demonstrate that by utilizing strategic conflict management methods, we can ensure a reliable foundation for effective tactical deconfliction for UAM. 
These complementary approaches work together to enhance the safety and efficiency of the UAM system.
\item{\bf Game theory to improve MARL convergence rate.} This paper focuses on analyzing the potential safety threats posed by multiple aircraft operating in close proximity, such as when two aircraft merge together. 
Specifically, we investigate the instability and convergence issues that arise when training a multi-agent reinforcement learning (MARL) model. 
Through our analysis, we identify the reasons behind the model's instability and introduce a new policy to mitigate this issue using game theory. 
Our numerical results demonstrate a significant improvement over the previous model, highlighting the effectiveness of our proposed approach.
\item{\bf A open-source UAM conflict mitigation sandbox.} We have made the code base of our integrated conflict management simulation, which utilizes the BlueSky simulator, publicly available. 
Our code includes baseline methods and evaluation metrics, enabling users to easily assess the performance of their own strategic and tactical algorithms by replacing the existing ones. 
This open framework allows for continued development and testing of conflict management approaches in the context of UAM, ultimately improving the safety and efficiency of the system.
\item{\bf Revealing essential insights into the interactions between strategic conflict management and tactical deconfliction.} 
In this paper, we demonstrate that strategic conflict management methods, such as departure separation and DCB, can effectively precondition tactical deconfliction and maintain safety metrics at nearly constant levels. 
In addition, tactical deconfliction methods improve traffic efficiency by permitting higher capacity near bottlenecks. 
However, the maneuvers employed by tactical deconfliction also result in demand uncertainty at each capacity constrained resource, which diminishes the effectiveness of DCB. 
\end{enumerate}

In Section \ref{formulation}, the problem formulation and system framework are described. 
In Section \ref{strategic}, we described the strategic conflict management methods, including departure separation and three different approaches for DCB. 
In Section \ref{tactical}, the multi-agent reinforcement learning separation method and a baseline method for tactical deconfliction are described. 
In Section \ref{exp}, five numerical experiments are described to demonstrate the effectiveness and interactions between strategic and tactical methods. 
Finally, we present conclusions in Section \ref{conclusion}.

\section{Problem Formulation} \label{formulation}

This paper aims to develop a system that ensures aviation safety metrics remain below target levels while optimizing traffic efficiency. 
To achieve this objective, we introduce an integrated conflict management platform (ICMP) for UAM, which integrates strategic and tactical separation methods to mitigate conflicts. 
As previously demonstrated in \cite{chen2022estimating}, a combination of strategic conflict management and tactical deconfliction is an effective approach for balancing safety and efficiency.

\subsection{Framework for Integrated Conflict Management Platform}
\begin{figure*}[ht!]
  \centering
  \includegraphics[width=0.8\textwidth]{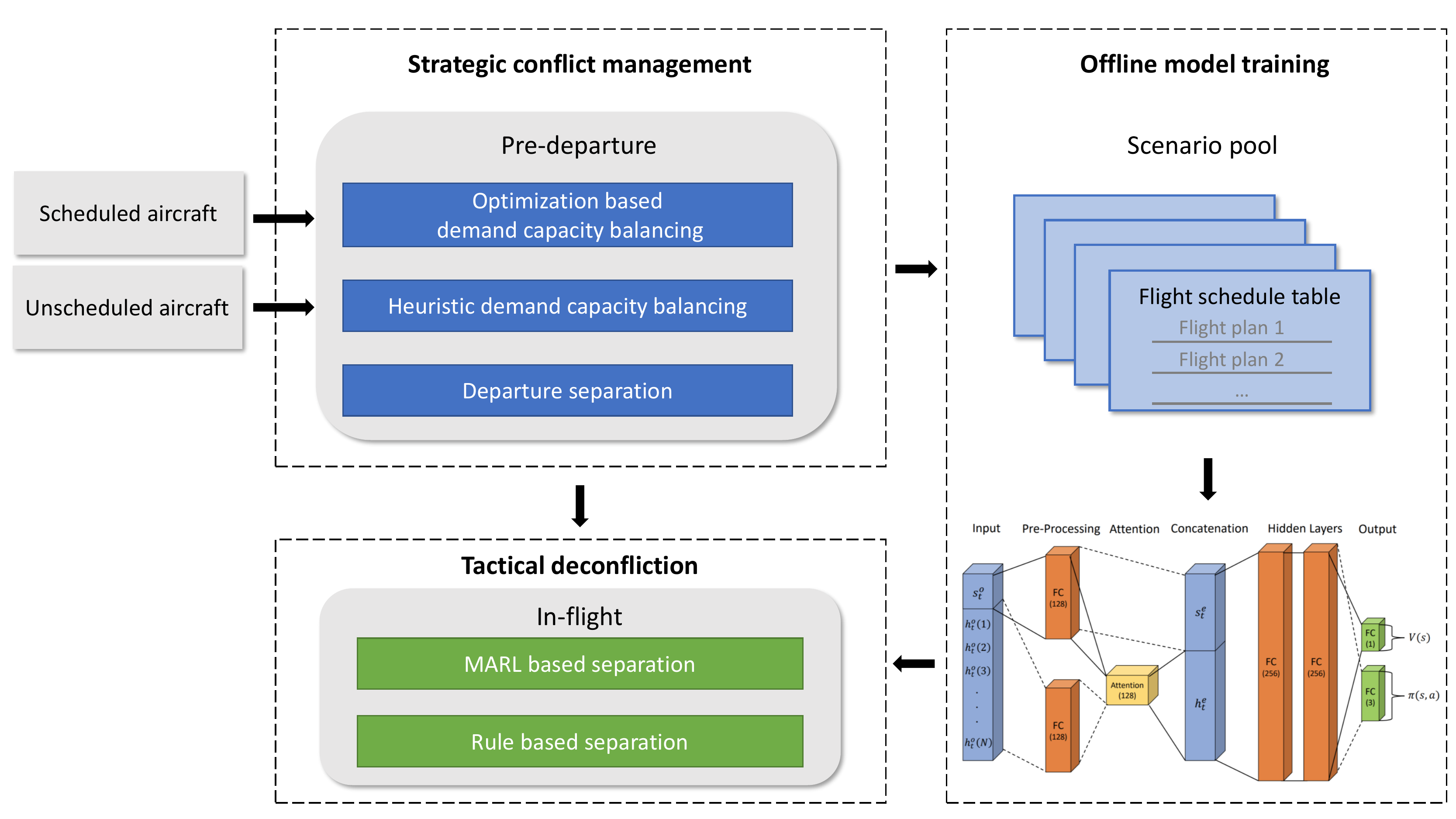}
  \caption{The framework of integrated conflict management platform.}
  \label{framework}
\end{figure*}
Figure \ref{framework} illustrates the ICMP framework, which divides the flight operation into two stages: pre-departure and airborne. 
The pre-departure stage utilizes strategic conflict management to determine an appropriate departure time by introducing ground delays. 
This paper presents one departure separation method and two demand capacity balancing algorithms to suit different scenarios, as outlined in Section \ref{strategic}. 
The airborne stage employs tactical deconfliction methods to provide speed advisories for all aircraft to resolve conflicts. 
This includes a MARL-based separation assurance method and a rule-based separation algorithm, the latter representing a benchmark against which the performance of the other methods can be compared. These tactical deconfliction methods are described in Section \ref{tactical}. 
Additionally, strategic conflict management generates simulated flight plans for the MARL offline model training process, which is then used for online operations.

\subsection{Safety Metrics}
In this paper, four safety metrics are measured: 
\begin{enumerate}
\item {\bf Number of Loss of Well Clear (LoWC) events per flight hour.} A LoWC event is defined as a loss of horizontal separation between any aircraft, and the range is set as 500 meters in this paper under the recommendation of \cite{weinert2018well}. 

\item {\bf Number of Near Mid Air Collisions (NMAC) per flight hour.}
Since Mid Air Collisions (MACs) between aircraft are rare, a Near Mid Air Collision (NMAC) is defined which represents a precursor to a Mid Air Collision. 
For crewed aviation, an NMAC is typically defined as a loss of 500 feet (152 meters) of horizontal separation and 100 feet (30 meters) of vertical separation\cite{weinert2020quantitatively}. 
Since we simulate operations flying at a co-altitude, we define an NMAC as a loss of 150 meters of horizontal separation, as described in Table \ref{Table mac}. 

\item{\bf Estimated Number of Mid Air Collisions (MAC) per flight hour.}
As described in Table \ref{Table mac}, we define a Mid Air Collision as a loss of horizontal separation of 10 meters, which is representative of the wingspan or maximum horizontal dimension of a UAM aircraft. 
However, since actual MACs are infrequent, especially with advanced conflict management, we instead observe the number of NMACs, and use a conditional probability, $\mathbb{P(\text{MAC}|\text{NMAC})}$, to estimate the probability of MAC. 
It's worth noting that this paper does not model the effect of collision avoidance systems such as the Airborne Collision Avoidance System X (ACAS X), which provides vertical and horizontal maneuvers to avoid mid-air collisions \cite{owen2019acas,alvarez2019acas}. 
Instead, we utilize a $\mathbb{P}(\text{MAC})$ risk ratio $\beta$ to compensate for the impact of airborne collision avoidance on the likelihood of a mid-air collision.

The ACAS X risk ratio $\beta$ is defined as:
% \begin{equation}
%     \beta =  \frac{\mathbb{P}(\text{MAC}|\text{NMAC, with ACAS})}{\mathbb{P}(\text{MAC}|\text{NMAC, without ACAS})}
% \end{equation}

\begin{equation}
    \beta =  \frac{\mathbb{P}(\text{NMAC, with ACAS X})}{\mathbb{P}(\text{NMAC, without ACAS X})}
\end{equation}

We estimate the number of MAC events $\mathcal{N}_{\text{MAC}}$ by:
% \begin{equation}
%  \mathbb{E}(\mathcal{N}_{\text{MAC}})=\mathbb{P}(\text{MAC}|\text{NMAC, with ACAS}) \cdot \mathcal{N}_{\text{NMAC}}    
% \end{equation}

\begin{equation}
 \mathbb{E}(\mathcal{N}_{\text{MAC}})=\mathbb{P}(\text{MAC}|\text{NMAC}) \cdot \beta \cdot \mathcal{N}_{\text{NMAC}}    
\end{equation}

where $\mathcal{N}_{\text{NMAC}}$ is the number of NMACs observed in the simulation without the implementation of ACAS X. 
The $\mathbb{P}(\text{MAC}|\text{NMAC})$ is obtained by using Monte Carlo simulation on a scenario without any intervention. 
Table \ref{Table mac} presents each of the parameter values used in the estimation of the number of MAC events.

\begin{table}[h]
\centering
\caption{simulation for estimated macs}
\label{Table mac}
\begin{tabular}{|c|c|}
\hline
MAC horizontal separation threshold & 10m \\ \hline
NMAC horizontal separation threshold & 150m    \\ \hline
Number of simulation runs (unmitigated)                    & 200    \\ \hline
Total testing flight hours (unmitigated)                     & 1000    \\ \hline
ACAS X risk ratio $\beta$\tablefootnote{In this paper, we choose $\beta$ as 0.005, which is calculated from \cite{katz2022collision} table 5, where the P(NMAC) without ACAS X = $3.01\times 10^{-3}$ and P(NMAC) with ACAS X vertical and speed advisories = $1.50\times 10^{-5}$.} & 0.005 \\ \hline
$P(\text{MAC}|\text{NMAC})$      & $5.038\times 10^{-3}$ \\ \hline
% $P(\text{MAC}|\text{NMAC, with ACAS})$      & $2.519\times 10^{-5}$ \\ \hline
\end{tabular}
\end{table}

\item{\bf Risk Ratio}
The risk ratio is calculated as the ratio of the number of estimated MACs for the non-intervention scenario and the number of estimated MACs for the other methods applying conflict management. 
\end{enumerate}

\subsection{Efficiency Metrics}
We calculate three different efficiency metrics:
\begin{enumerate}
\item {\bf Ground delay} due to strategic conflict management. 
If departure demand is sufficiently high that demand exceeds capacity for any constrained resources, DCB will calculate a new departure time for the aircraft that will prevent the demand from exceeding capacity. 
Ground delay is calculated as:

\begin{equation}
\text{ground delay} = \max\{0, (R_f - S_f)\}    
\end{equation}

where $R_f$ is the required departure time of flight $f$ and $S_f$ is the original scheduled departure time.

\item {\bf Airborne delay} due to tactical deconfliction. 
For each aircraft, we estimate total flying time $T_{f}$ based on the distance and the aircraft cruise speed. 
During simulation, we implement tactical deconfliction and measure the actual flying time $A_f$. 
Airborne delay is calculated as follows:
\begin{equation}
    \text{airborne delay}=\max\{0, (A_f-T_f)\}
\end{equation}

\item {\bf Number of alerts} is the total number of speed-change advisories requested by the tactical deconfliction methods. 
Operators generally seek to minimize the number of maneuvers in the air, which use increased energy and increase workload on pilots. 
Hence, the number of alerts is applied as an efficiency metric.
\end{enumerate} 

\section{Strategic Conflict Management} \label{strategic}

DCB is a mechanism that has been identified by the Federal Aviation Administration (FAA) as potentially being required to support urban air mobility (UAM) operations as the number of operations increases \cite{faa2020uam}. 
DCB involves defining airspace capacity and managing demand strategically to prevent demand from exceeding capacity. 
This can help to balance efficiency and predictability in UAM operations, particularly when operational uncertainties are high. 
By using DCB, it is possible to manage the demand for constrained resources, such as airspace network intersection points, in a way that helps to ensure the smooth and safe operation of UAM vehicles.

\begin{figure}[ht]
% \vskip 0.2in
\begin{center}
\centerline{\includegraphics[width=\columnwidth]{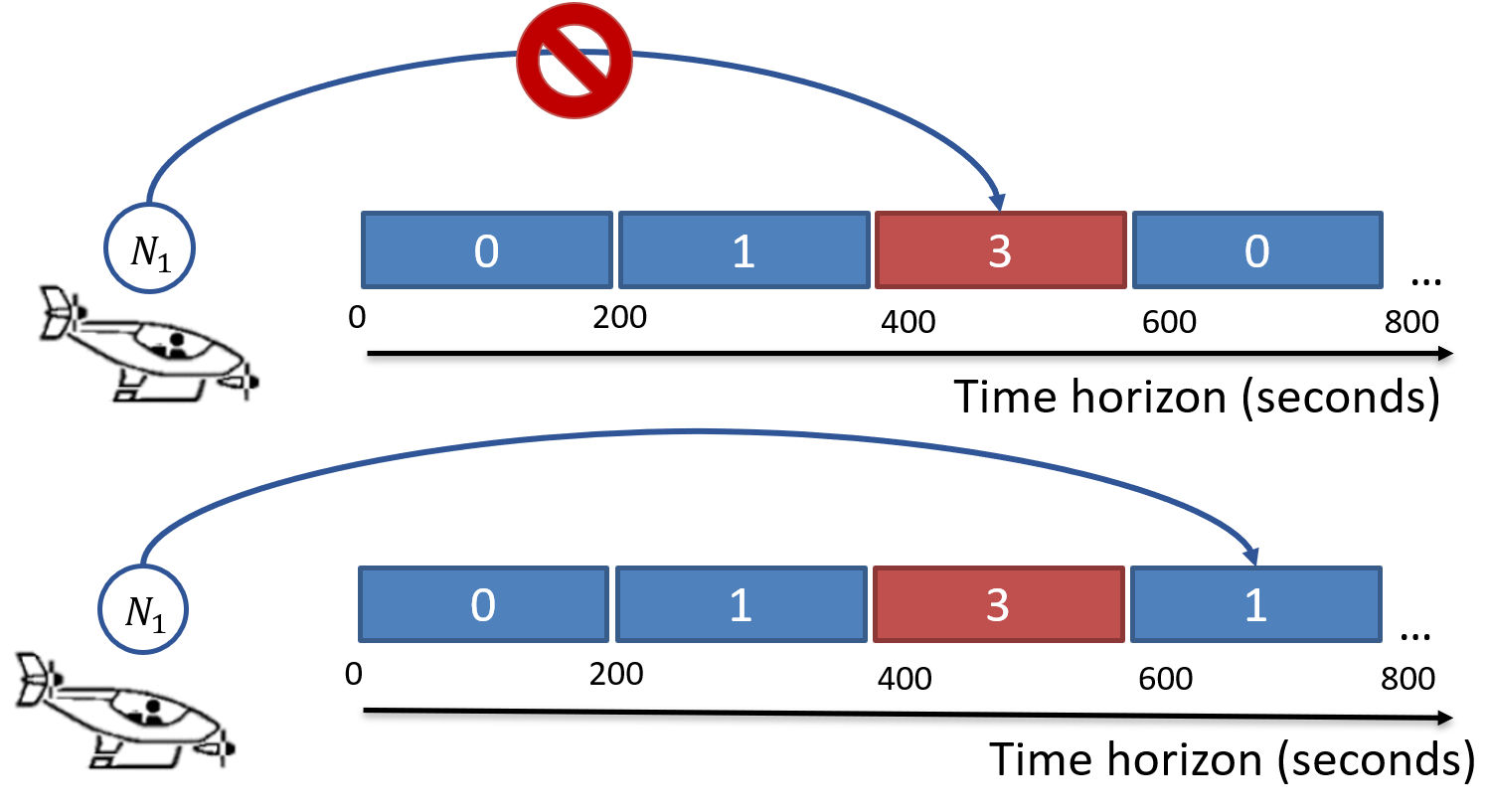}}
\caption{Diagram of DCB. For the given bottleneck with a capacity of 3 operations every time window of 200 seconds, the aircraft's estimated arrival time falls within a fully occupied time window. 
To ensure the aircraft's arrival at the next available time window, the operation is assigned a ground delay.}
\label{DCB}
\end{center}
\vskip -0.2in
\end{figure}

Figure \ref{DCB} shows a schematic diagram of the DCB algorithm. 
At each bottleneck (crossing or merge point), the time horizon is divided into multiple time windows, each with a fixed duration $S$. 
The capacity $C$ of the resource defines the maximum number of flights that can fly through the resource in the same time window. 
The goal of DCB is to strategically control the throughput at each bottleneck by delaying operations on the ground.

\begin{figure*}[t]
\centering
\subfloat[]{\includegraphics[width=\columnwidth]{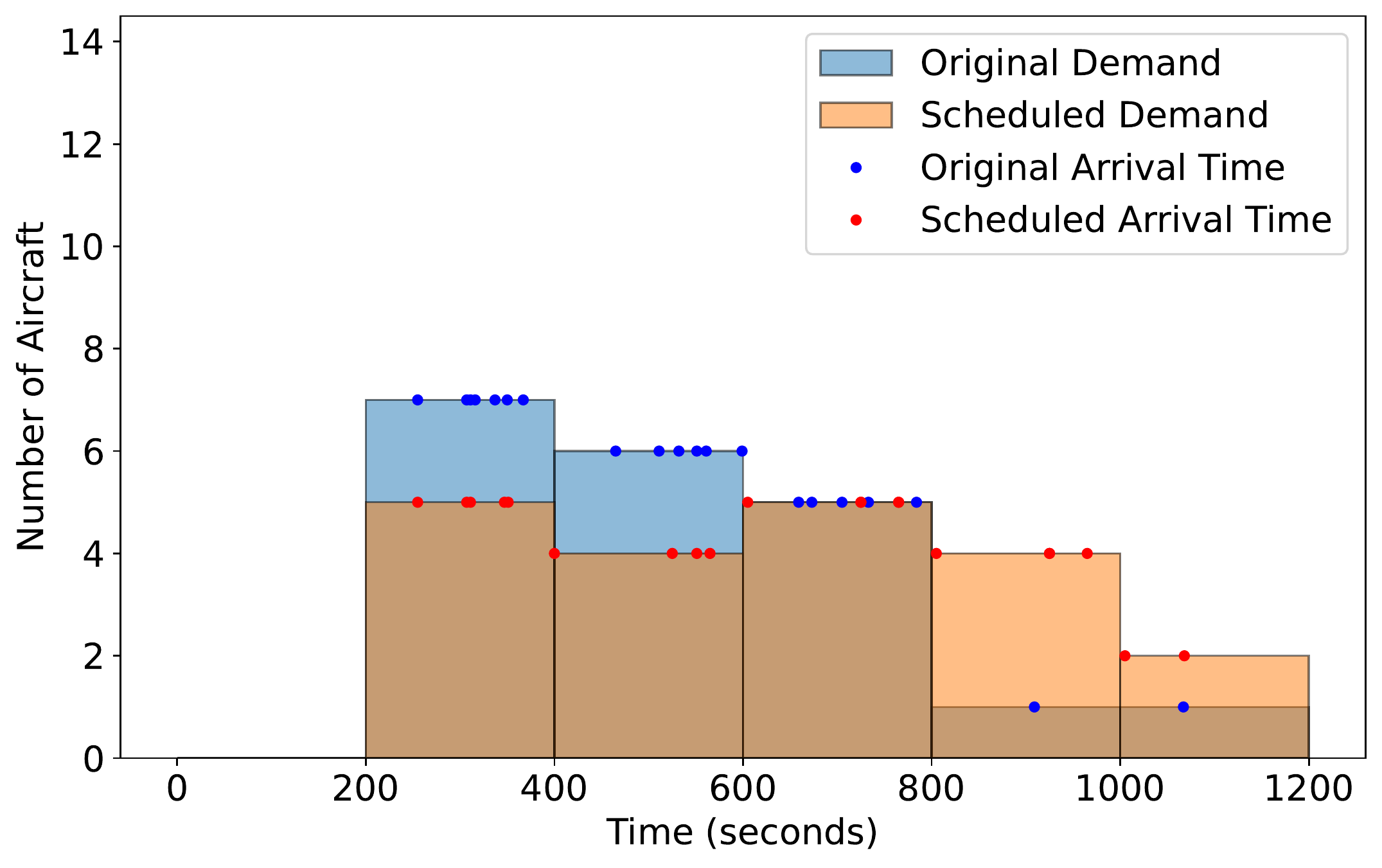}%
\label{fig_first_case}}
\hfil
\subfloat[]{\includegraphics[width=\columnwidth]{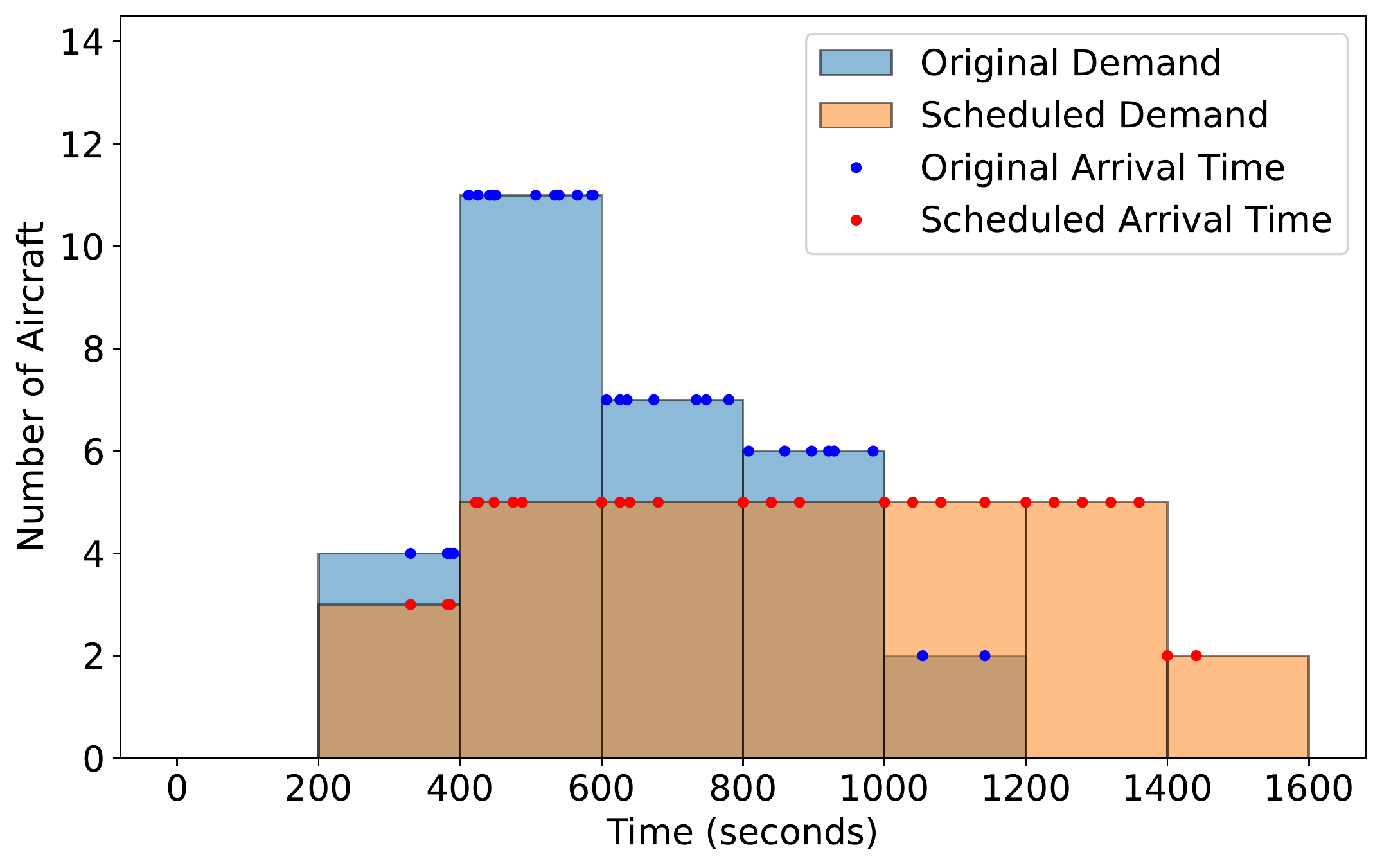}%
\label{fig_second_case}}
\caption{Example of how DCB can be applied across multiple resources. The blue bars show the original demand across different time windows, and the orange bars show the optimized traffic demand. The blue and orange dots are the exact departure times of the modeled operations. (a) Traffic demand on resource 1. (b) Traffic demand on resource 2.}
\label{DCB_PERF}
\end{figure*}

This paper introduces two DCB algorithms with different applications. 
An optimization-based DCB algorithm is centralized and takes the scheduled departure time of all aircraft as input and calculates the optimal departure time required to minimize total delay in advance. 
In contrast, the heuristic DCB algorithm has no guarantee to minimize total departure delay but can be used to determine ground delays required to ensure that demand does not exceed capacity for unscheduled aircraft in real-time. 
While optimization-based DCB works well for scheduled demand, heuristic DCB is useful for inserting unscheduled demand into the calculated departure flow. Figure \ref{DCB_PERF} provides an example of how DCB works across multiple resources.

\subsection{Optimization Based Demand Capacity Balancing}

In this problem, we formulated the DCB problem into a mix-integer programming problem, which can solve for networks with multiple capacity constrained resources. 
The formulation is shown below.

\begin{align}
\min_{\boldsymbol{\omega}\in \mathds{B}^+, \boldsymbol{R}\in \mathds{R}^+}&   \sum_{d\in \mathcal{D}}\sum_{f \in \mathcal{F}^d}(R_{d,f}-S_{d,f})\label{obj}\\
\textrm{s.t.} \quad &R_{d,f+1}-R_{d,f} \ge \Delta, \hspace{0.35in}\forall d\in \mathcal{D},  f \in \mathcal{F}^d\label{c1}\\
&R_{d,f} \ge S_{d,f}, \hspace{0.8in}\forall  d\in \mathcal{D}, f \in \mathcal{F}^d\label{c2}\\
&\sum_{n\in \mathcal{N}} \omega_{n, d,f, i} = 1,\quad \forall d\in \mathcal{D}, f \in\mathcal{F}^d, i\in\mathcal{I}^d \label{c3}\\
&(R_{d, f}+T_{d, i}-B_n)\omega_{n, d,f, i} \ge 0, \label{c4}\\
&\hspace{0.65in}\forall d\in \mathcal{D}, f \in\mathcal{F}^d,i\in\mathcal{I}^d, n \in \mathcal{N}\notag\\
&(R_{d, f}+T_{d, i}-B_n)\omega_{n, d, f, i} \le W, \label{c5}\\
&\hspace{0.65in}\forall d\in \mathcal{D}, f \in\mathcal{F}^d,i\in\mathcal{I}^d, n \in \mathcal{N}\notag\\
&\sum_{d\in \mathcal{D}}\sum_{f \in \mathcal{F}^d}\sum_{i \in \mathcal{I}^d, i=p}\omega_{n, d, f, i} \le C^p,\label{c6}\\
&\hspace{1.55in}\forall n \in\mathcal{N},p\in\mathcal{P}\notag
\end{align}
 
In this formulation, two decision variables are introduced: the time window identifier $\omega$, and the required time of departure $R$. 
The objective (equation (\ref{obj})) of the problem is to minimize the ground delay of all aircraft $f \in \mathcal{F}^d$ on all routes $d \in \mathcal{D}$. 
Here, $R_{s,d}$ is the required time of departure, and $S_{d,f}$ is the original scheduled departure time.

Constraint (\ref{c1}) ensures that any two aircraft departing from the same vertiport have a minimum separation of $\Delta$. 
Constraint (\ref{c2}) ensures that the required departure time is not earlier than the scheduled time. 
Constraints (\ref{c3})-(\ref{c5}) are used to identify the time window to which the estimated arrival time belongs. 
Here, $\omega_{n,d,f,i}=1$ means that aircraft $f$ departing from $d$ will arrive at resource $i$ during time window $n$. 
$R_{d,f}+T_{d,i}-B_n$ is the relative arrival time compared to time window $n$, where $T_{d,i}$ is the estimated flying time from $d$ to $i$, $B_n$ is the start time of time window $n$, and $W$ is the length of the time window (set to 200 seconds in this paper). 
The identifier is activated as 1 only when the relative arrival time is within the interval $[0, W]$, and it can only be activated once. 
Finally, constraint (\ref{c6}) ensures that the number of aircraft at each resource $p\in \mathcal{P}$ does not exceed the capacity $C^p$ of the resource. 
It is worth noting that resource set $\mathcal{I}^d$ includes only the resources involved in the route starting from $d$, while resource set $\mathcal{P}$ includes all the actual capacity constrained resources in the airspace. 

\subsection{Heuristic Demand Capacity Balancing}

A single resource heuristic DCB algorithm is proposed in \cite{chen2022estimating}. 

\begin{algorithm}[H]
\caption{Heuristic Demand Capacity Balancing}\label{alg:DCB}
\begin{algorithmic}
\STATE
\STATE Collect initial DCB window list $\boldsymbol{\omega}$
\STATE Initialize start time $t$
\STATE \textbf{while} {$t < T$}:
\STATE \hspace{0.5cm}  BlueSky.step()
\STATE \hspace{0.5cm}  $t += \text{SIMDT}$
\STATE \hspace{0.5cm} \textbf{if} {received departure request from aircraft f at route r}:
\STATE \hspace{1cm} check departure time of ahead aircraft $R_{r, f-1}$
\STATE \hspace{1cm}\textbf{if} $(R_{r, f}-R_{r, f-1})\ge \Delta$:
\STATE \hspace{1.5cm}\textbf{if} $\boldsymbol{\omega}.\text{map}(t+D_{i}) < C_i$ for all bottlenecks:
\STATE \hspace{2.0cm}Release aircraft $f$
\STATE \hspace{2.0cm}$\boldsymbol{\omega}.\text{map}(t+D_{i})+=1$
\end{algorithmic}
\label{alg1}
\end{algorithm}

In our paper, we improved the algorithm to support networks with multiple resources. 
When the system receives new demand for the resource, it first checks the departure time of the leading aircraft. 
If the departure separation is within the required separation $\Delta$, the system then uses a mapping function to check the remaining volume of the corresponding window. 
If the demand in the window reaches any of the involved resource capacities $C_i$, the following departure will be prevented from departing until the next window that is under the capacity limit appears. 
This algorithm is detailed in Algorithm \ref{alg1}.

\section{Tactical Deconfliction}\label{tactical}
As introduced in \cite{chen2022estimating}, strategic deconfliction can mitigate conflicts and guarantee safe separation but at a significant cost to efficiency. 
To enhance safety and efficiency under uncertainty, airborne operations require tactical deconfliction, which provides maneuver advisories to resolve potential conflicts. 
In this paper, we introduce two tactical deconfliction methods, i.e., a learning-based method, and a rule-based method.
\subsection{MARL Tactical Deconfliction}\label{reward function}
The multi-agent reinforcement learning (MARL) algorithm to control individual aircraft in a simulated air traffic environment is originally introduced in \cite{brittain2019autonomous} and improved in \cite{D2MAV-A}. 
By using MARL, the algorithm can adapt to changing conditions and learn from past experience, which can help to improve the performance of the system over time. 
Additionally, by training all of the agents using a shared model, all of the aircraft are following the same separation policy, which can help to prevent conflicts and maintain a safe and efficient flow of traffic. 
Overall, this approach combines the advantages of MARL and shared model training to provide a powerful tool for aircraft tactical deconfliction.

The MARL model is formulated as follows based on \cite{D2MAV-A}:

\subsubsection{\bf State Space}
In reinforcement learning, the state space refers to the set of all possible states that an agent can encounter at a given time. 
In this particular study, we assume that the aircraft's state and dynamics information is fully accessible to others, like position, speed, and distance to the destination. 
Specifically, the state space for each agent is formulated as follows:
\begin{align}
s^o_t = \{ d_{goal}^{(o)}, v^{(o)} \theta^{(o)},  d_{\text{NMAC}}\}\\
h_t^o(i) = \{ d_{goal}^{(i)}, v^{(i)},  \theta^{(i)}, d_o^{(i)}\}
\end{align}
where $s^o_t$ represents the state of the ownship, which contains the distance to the goal $d_{goal}^{(o)}$, aircraft speed $v^{(o)}$, aircraft heading $\theta^{(o)}$, and the NMAC boundary $d_{\text{NMAC}}$. 
The state of the intruder is quite similar to the ownship while replacing the NMAC boundary with the distance between the ownship and the intruder $d_o^{(i)}$.
\subsubsection{\bf Action Space}
In this study, the action space is defined as the set of possible actions that an aircraft can take at each decision-making step. 
These actions include decreasing speed, holding the current speed, or increasing speed:
\begin{equation}
  A_t = [-\Delta v, 0, +\Delta v]  
\end{equation}

\subsubsection{\bf Reward Function}
A reward function in reinforcement learning can provide a scalar feedback signal to an agent, indicating the desirability of the state-action pair taken by the agent in an environment. 
In this paper, three types of penalties are considered:  
\begin{equation}
R(s,t,a)=R(s)+R(t)+R(a)
\end{equation}

Since maintaining separation is the primary objective in this paper, the majority of the reward function during the training process is allocated to the safety penalty term, denoted as $R(s)$. 
The safety penalty is defined as follows:
\begin{equation}
    R(s)=\left\{
\begin{array}{lr}
-1 &\text{if}\quad d_o^{(i)} < d_{\text{NMAC}} \\
-\alpha+\delta\cdot d_o^{(i)}  &\text{if}\quad d_{\text{NMAC}}\le d_o^{(i)} \le d_{\text{LoWC}}  \\
0 &\text{otherwise}
\end{array}
\right.
\end{equation}
If the distance between the ownship and the intruder is within the NMAC threshold $d_{\text{NMAC}}$, the agent incurs a penalty of -1. 
If the distance falls between the NMAC threshold and the LoWC threshold, the penalty is linearly proportional to the distance.

The objective of this paper is also to enhance traffic efficiency while maintaining a specified level of safety. 
To achieve this goal, the second component of the reward function is the flying time penalty, denoted as $R(t)$.
\begin{equation}
    R(t)=\left\{
\begin{array}{lr}
-1 &\text{if}\quad t > T \\
-\eta &\text{otherwise}
\end{array}
\right.
\end{equation}
If an aircraft exceeds its maximum flying time $T$ and fails to reach its destination, it incurs a penalty of -1 and is removed from the simulation. 
Otherwise, a fixed penalty $\eta$ is applied at each step and accumulated over time. 
This encourages the agent to avoid local optima, where all aircraft maintain minimum speed until the end of the simulation, by increasing their speeds.

In the real world, frequent speed changes can increase pilot workload (in the case of a piloted aircraft), with associated safety implications, and can also result in higher energy use. 
To mitigate these risks, we introduce the action penalty term $R(a)$
\begin{equation}
    R(a)=\left\{
\begin{array}{lr}
0 &\text{if a = 0} \\
-\psi &\text{otherwise}
\end{array}
\right.
\end{equation}
Whenever an aircraft changes its speed, a fixed penalty $\psi$ is applied and accumulated over time. 
This penalty is intended to discourage unnecessary speed changes and encourage smoother flight paths.

\subsection{Implementation of Game Theory}
\begin{figure}[ht]
% \vskip 0.2in
\begin{center}
\centerline{\includegraphics[width=\columnwidth]{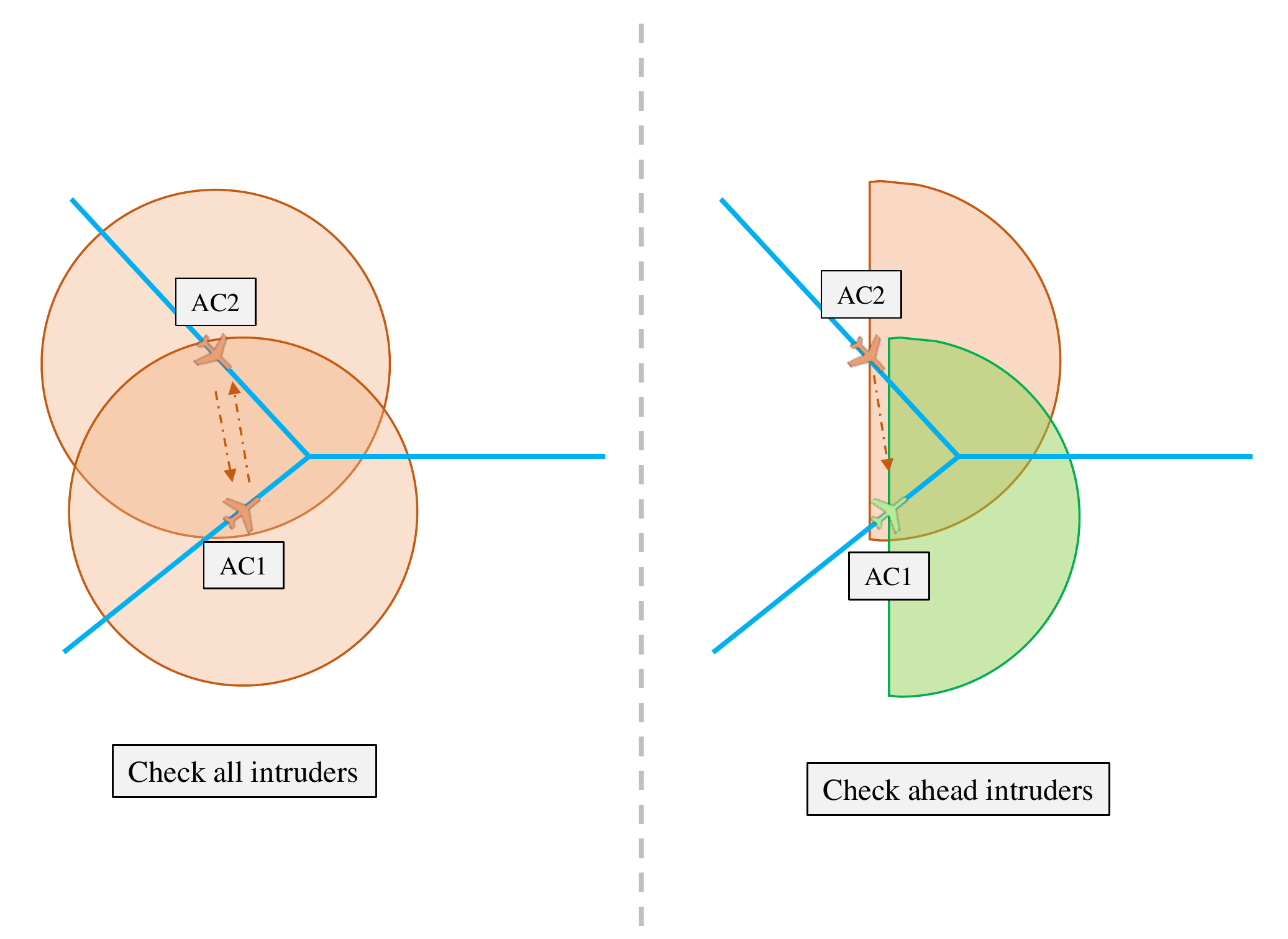}}
\caption{Different intruder detection policies.}
\label{game}
\end{center}
% \vskip -0.2in
\end{figure}

To gain a comprehensive understanding of multi-agent decision-making problems and enhance the efficacy of tactical deconfliction methods, it is valuable to analyze the relationships among agents. 
However, solving a detailed multi-stage decision-making problem for all the aircraft from start to end becomes challenging when applying game theory.
The equilibrium is hard to reach because of the computation complexity and inefficiency.
To break down the problem, we focus on a one-step decision-making scenario between two merging aircraft, as illustrated in Figure \ref{game}.
\begin{table}[h]
\caption{cost table}
\label{cost table}
\begin{tabular}{lllll}
                                          &                                 & \multicolumn{3}{c}{Aircraft 2}                                                                       \\ \cline{2-5} 
\multicolumn{1}{l|}{}                     & \multicolumn{1}{l|}{}           & \multicolumn{1}{l|}{Speed up} & \multicolumn{1}{l|}{Hold}   & \multicolumn{1}{l|}{Slow down} \\ \cline{2-5} 
\multicolumn{1}{l|}{\multirow{3}{*}{Aircraft 1}} & \multicolumn{1}{l|}{Speed up}   & \multicolumn{1}{l|}{-1, -1}   & \multicolumn{1}{l|}{-1, -1} & \multicolumn{1}{l|}{0, -0.01}   \\ \cline{2-5} 
\multicolumn{1}{l|}{}                     & \multicolumn{1}{l|}{Hold}       & \multicolumn{1}{l|}{-1, -1}   & \multicolumn{1}{l|}{-1, -1} & \multicolumn{1}{l|}{-1, -1}     \\ \cline{2-5} 
\multicolumn{1}{l|}{}                     & \multicolumn{1}{l|}{Slow down} & \multicolumn{1}{l|}{-0.01, 0} & \multicolumn{1}{l|}{-1, -1} & \multicolumn{1}{l|}{-1, -1}     \\ \cline{2-5} 
\end{tabular}
\end{table}

\begin{figure*}[!t]
\centering
\subfloat[]{\includegraphics[width=\columnwidth]{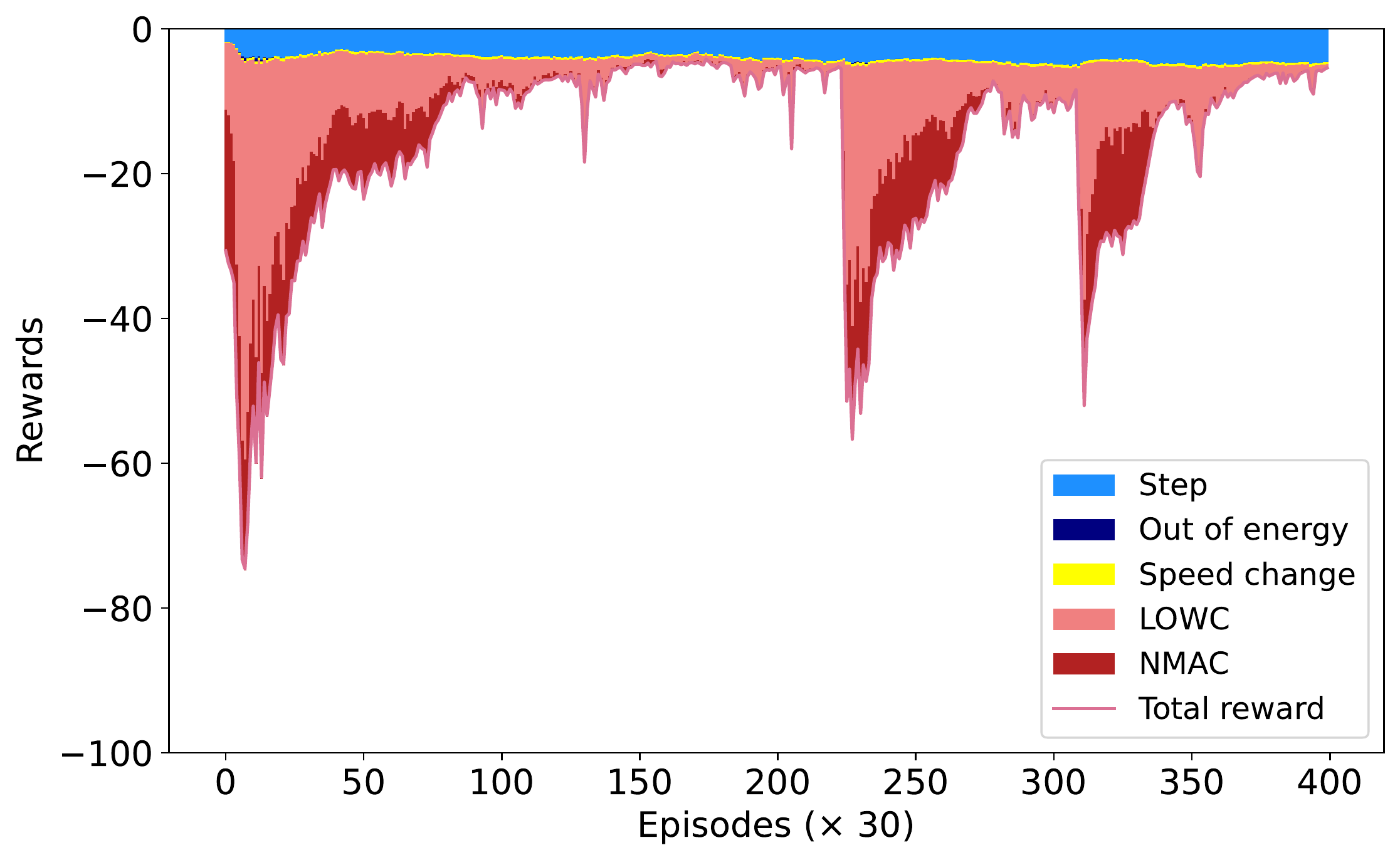}%
\label{fig_first_case}}
\hfil
\subfloat[]{\includegraphics[width=\columnwidth]{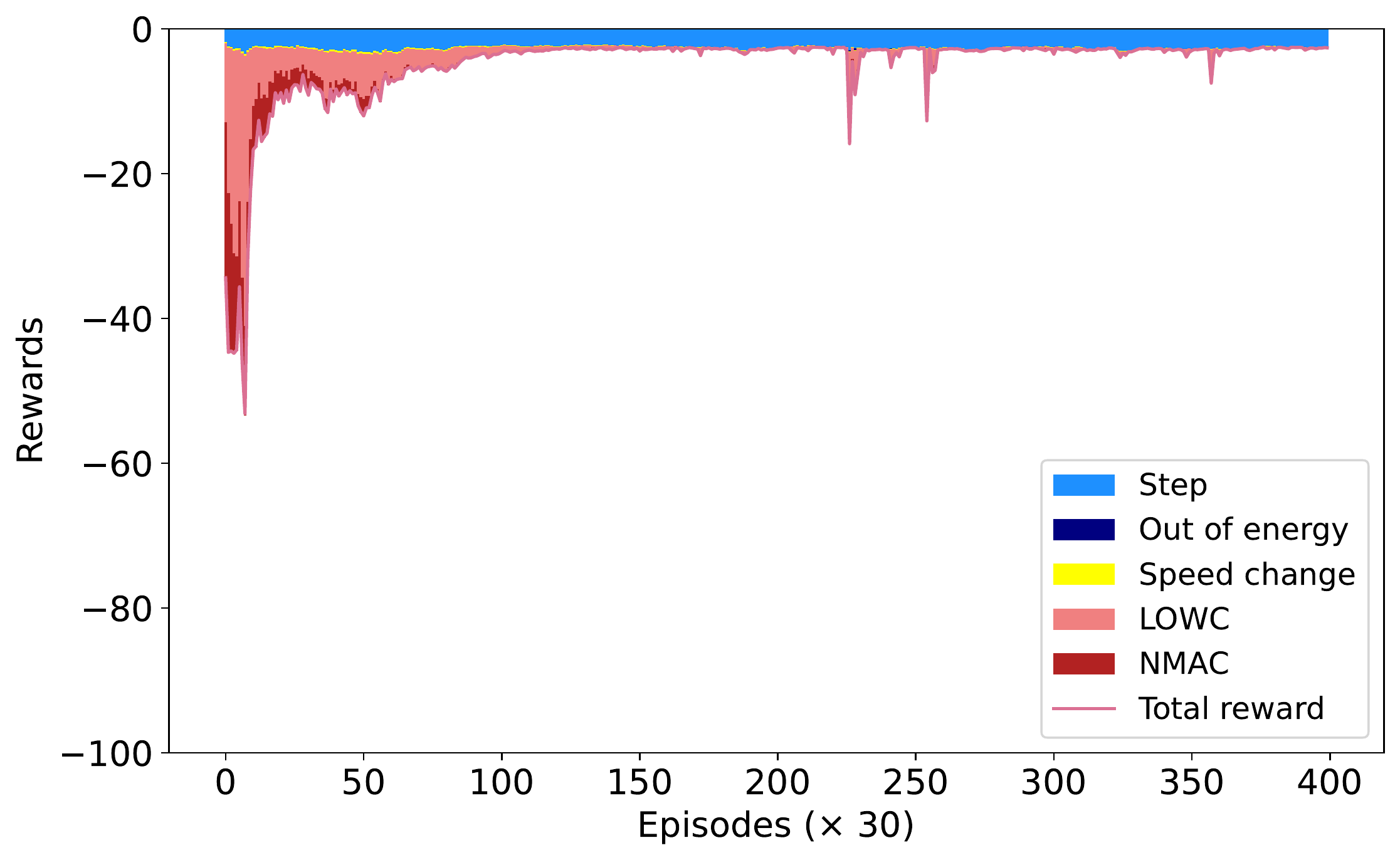}%
\label{fig_second_case}}
\caption{The learning curve of MARL with different intruder detection policies. (a)Detect all intruders nearby; (b) Only detect forward intruders.}
\label{game_learning}
\end{figure*}

The cost matrix of two aircraft on merging trajectories can be abstracted to that shown in Table \ref{cost table}. 
In this situation, only one speed decrease and one speed increase action can effectively mitigate the conflict, while any other actions may result in a significant penalty (-1). 
Additionally, choosing to decrease speed incurs a small additional energy cost (-0.01). 
In the previous work's setting \cite{brittain2019autonomous,D2MAV-A}, when Aircraft 1 and Aircraft 2 identify each other as intruders, the case is a general sum game with two equilibriums ([speed up, slow down], [slow down, speed up]). 
This ambiguous relationship leads to difficult decision-making for both aircraft, resulting in a lower convergence rate for multi-agent reinforcement learning (MARL) training. 
However, if a policy is implemented where aircraft only check for leading aircraft and make decisions in order, the case can be changed to a Stackelberg game, with only one dominant equilibrium ([speed up, slow down]). 
This new relationship is simpler, making it easier for agents to select the correct actions. 
Figure \ref{game_learning} shows the learning curve for MARL with different intruder detection policies.

\subsection{Rule-based Tactical Deconfliction}
\begin{figure}[h!]
\centerline{\includegraphics[width=\columnwidth]{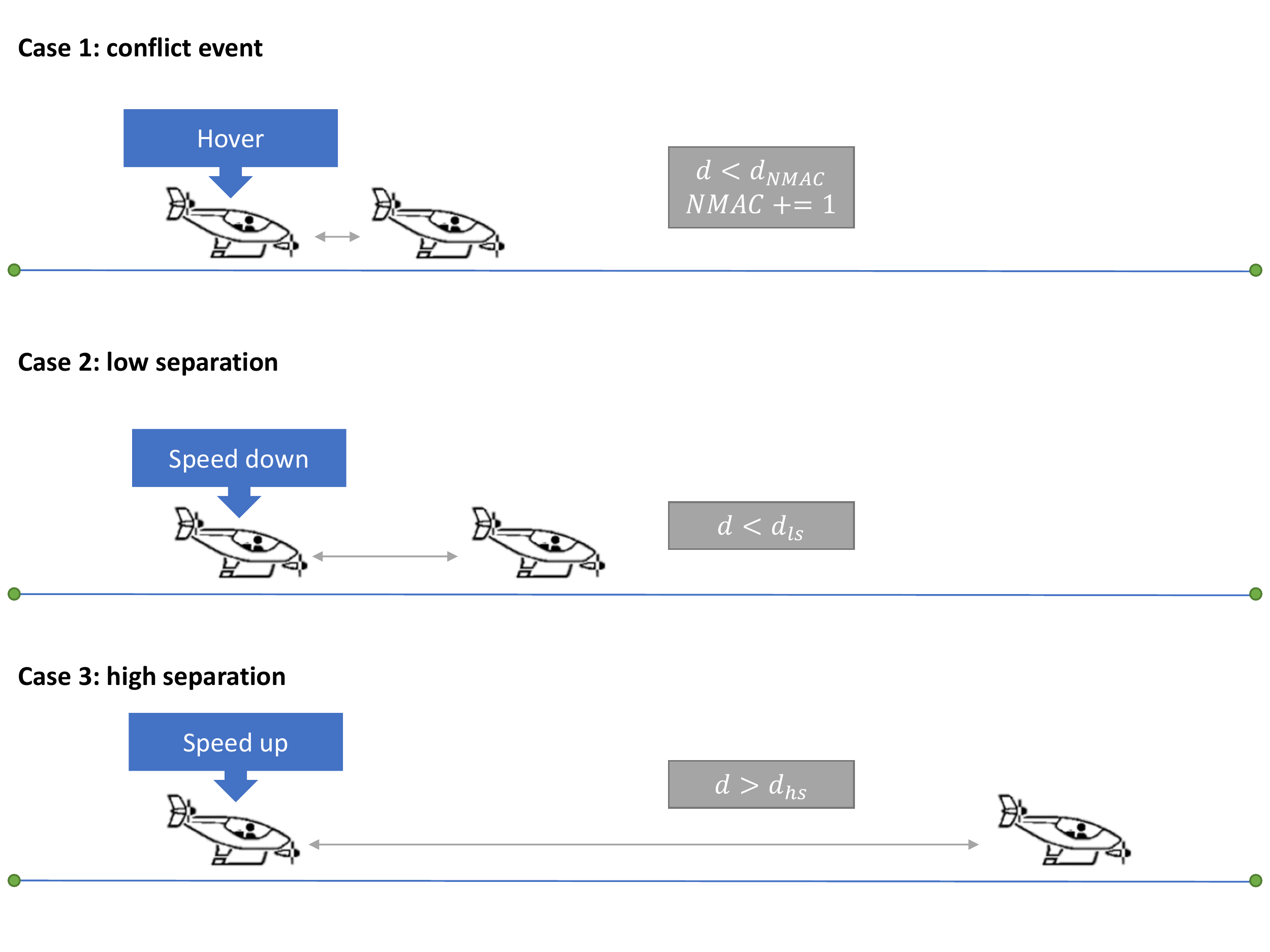}}
\caption{Description of rule-based tactical deconfliction.}
\label{rule-based}
\end{figure}

The rule-based tactical deconfliction method relies on predefined rules to determine the actions of aircraft to avoid NMACs, which is described in Figure \ref{rule-based}. 
The rules are based on specific thresholds for distances between aircraft, including the NMAC threshold $d_{\text{NMAC}}$, low separation boundary $d_{ls}$, and high separation boundary $d_{hs}$.

In the case where the distance between two aircraft is closer than the NMAC threshold, the following aircraft will choose to hover or reduce speed to a minimum level to avoid a potential collision. 
This situation is defined as an NMAC event. 
If the distance between the following aircraft and the lead aircraft is lower than the low separation boundary, the following aircraft will choose to slow down. 
On the other hand, if the distance is larger than the high separation boundary, or if there is no leading aircraft, the following aircraft will choose to speed up.

The rule-based tactical deconfliction method serves as a benchmark in the study described in \cite{chen2022estimating}. 
However, it should be noted that rule-based methods may have limitations in complex and dynamic environments, and it only provides a baseline approach for separation assurance but may require further refinement and improvement for more complex situations.

\section{Numerical Experiments}\label{exp}

\subsection{Simulation Environment}
In this study, we use BlueSky \cite{hoekstra2016bluesky} as our simulator to run a fast-time simulation. 
The BlueSky simulator is capable of running a large number of aircraft simulations in parallel efficiently.
In addition, it is highly configurable, e.g., allowing the configuration of vertiport locations, waypoint locations, UAM routes, and aircraft performance parameters. 

\begin{figure}[ht]
\begin{center}
\centerline{\includegraphics[width=\columnwidth]{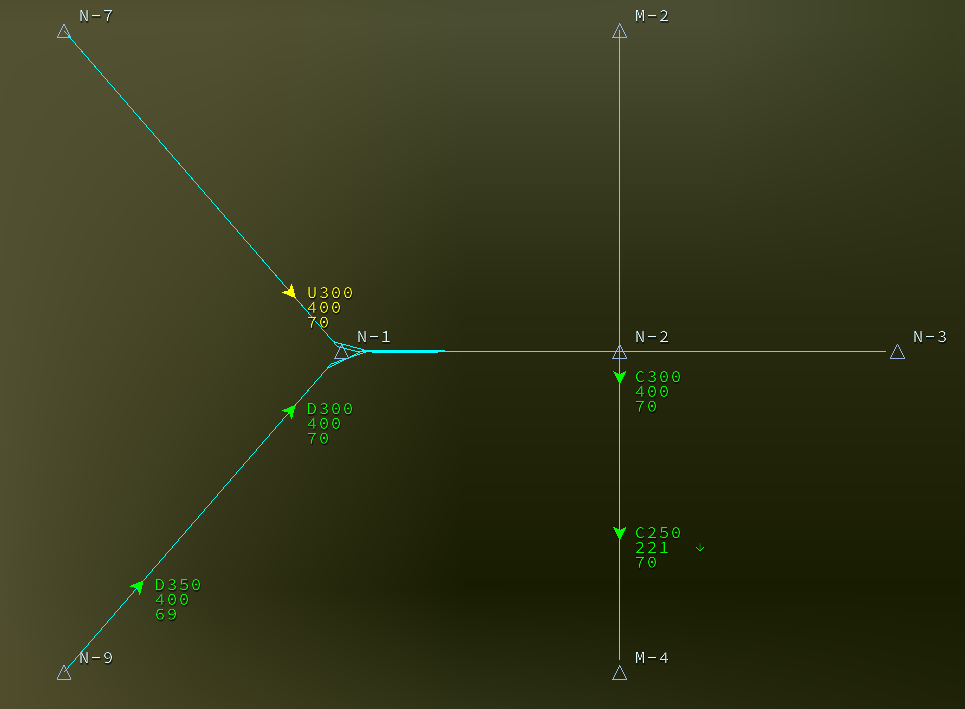}}
\caption{Illustration of the hybrid scenario in the Bluesky simulator. The blue lines represent three routes, while the grey triangles with labels indicate the origin point, destination, and waypoints. Additionally, the green and yellow triangles represent each aircraft, displaying information such as the aircraft ID, altitude, and speed.}
\label{scn}
\end{center}
\end{figure}

To study and evaluate the performance of the integrated conflict management system in structured airspace, we develop an evaluation scenario as shown in Figure \ref{scn}, which defines capacity constrained resources as the typical bottlenecks in an airspace network. Three routes are included in the scenario:
\begin{itemize}
    \item N-7 $\rightarrow$ N-1 $\rightarrow$ N-2 $\rightarrow$ N-3
    \item N-9 $\rightarrow$ N-1 $\rightarrow$ N-2 $\rightarrow$ N-3
    \item M-2 $\rightarrow$ N-2 $\rightarrow$ M-4
\end{itemize}
where N-1 and N-2 are two resources in this network. We implement the optimization-based DCB on both resources. 

\subsection{Experimental Results}
We conducted several numerical experiments to showcase the efficacy of our proposed integrated conflict management framework. 
Specifically, we first demonstrate the learning curve of the MARL training process on different capacities, which is used to determine the proper traffic density to obtain the best MARL model. 
Next, we determine the maximum capacity for the rule-based tactical method and MARL deconfliction model, as well as the capacity without any tactical deconfliction as a reference. 
Once we have the proper capacity value for DCB, we use those determined parameters and the trained model to compare the performance of different algorithm combinations using six metrics. 
Finally, we analyze the speed curve of various tactical deconfliction methods to gain insights into the reasons for differences in performance. 

To ensure a fair comparison between the rule-based and MARL methods, we set the observation range to 1500m for both methods. 
The decision-making of the ownship aircraft would be affected by the intruders who are within this observation range.

\subsubsection{\bf Learning Curve for Different Capacities}

The ultimate goal of the MARL model is to reduce penalties and determine the best policy for a given environment. 
However, if the traffic density is too high, or if aircraft do not have sufficient initial separation, it can be challenging for the MARL model to search for the optimal policy. 
In fact, high traffic density may lead the model to an unexpected local optimal policy, such as forcing all aircraft to airborne holding to avoid conflicts or even colliding to avoid further penalty steps. 
Therefore, it is essential to have a DCB layer as a precondition for MARL training.

To train the MARL model, we utilized the flight schedule tables optimized by DCB as the training scenario. 
To generate these tables, we first created a set of original scheduled departure times $S_{d, f}, \forall f \in F^d$, corresponding to three departure points $d\in D$, independently. 
The departure intervals $S_{d, f+1}-S_{d, f}$ on each route follow a beta distribution, with the average interval $\lambda$ used to control the traffic demand.
Next, we used DCB with a fixed capacity to compute the set of required departure times $R_{d,f}, \forall d\in D, f\in F^d$, and compiled them into a flight schedule table.
Each table contains 30 flight plans, which include information such as required departure time, origin, destination, waypoints, cruise speed, and cruise altitude.
To avoid overfitting, we generated 100 different flight schedule tables and place them into a scenario pool.
During training, the MARL model randomly selected a flight schedule table at each episode to improve its generalization performance. 
An episode is defined as a simulation round that fully executes the flight schedule table, starting from the first aircraft departure and ending with the final aircraft landing.

The training process consisted of a total of 150,000 episodes and was performed on two Nvidia RTX 3090 graphics cards. 
The model updated its weight every 30 episodes, and the simulation was executed in parallel with the support of the Ray python package~\cite{ray}. 
The entire training process took roughly 4 hours. 

\begin{figure*}[!t]
\centering
\subfloat[]{\includegraphics[width=\columnwidth]{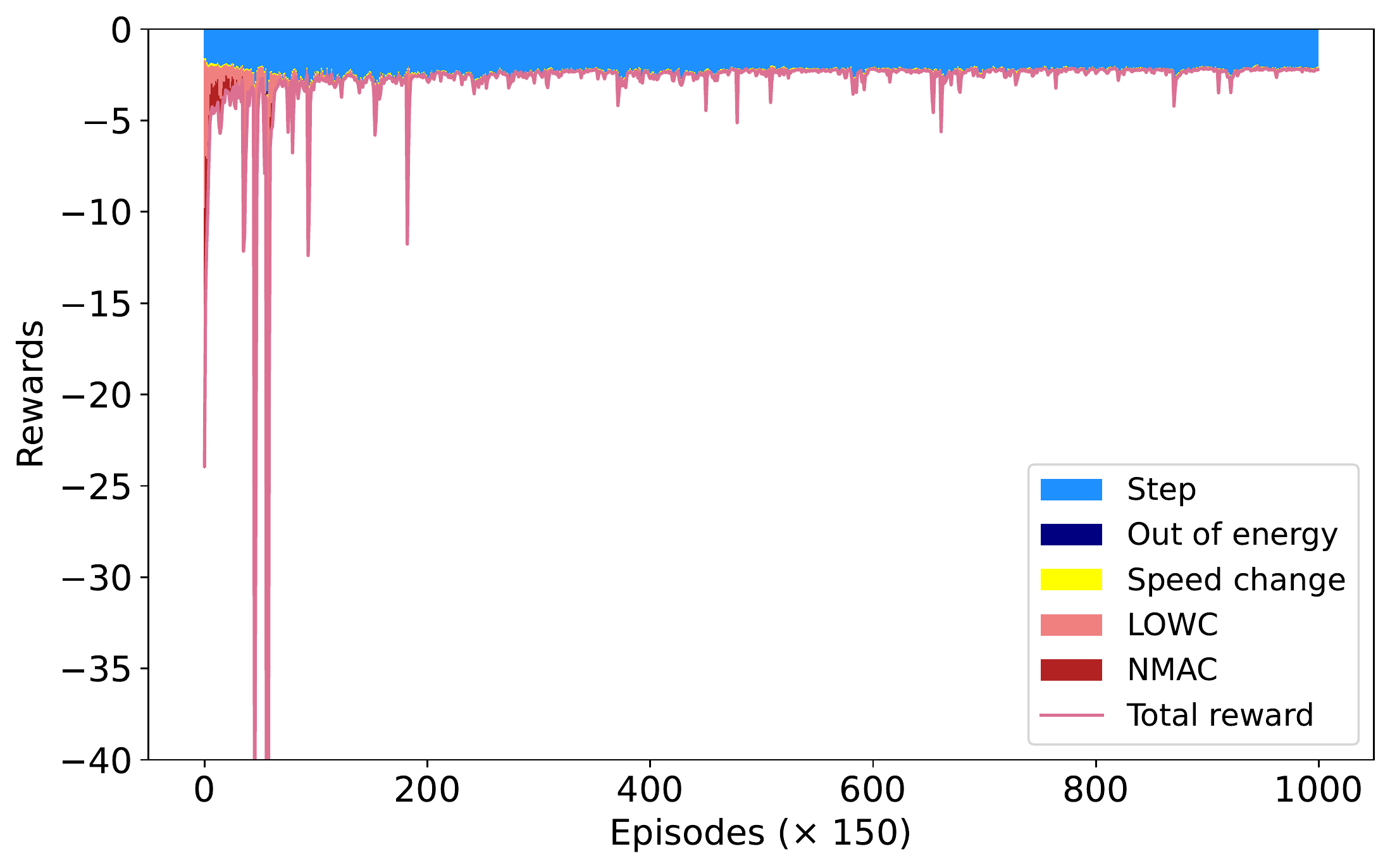}%
\label{fig_first_case}}
\hfil
\subfloat[]{\includegraphics[width=\columnwidth]{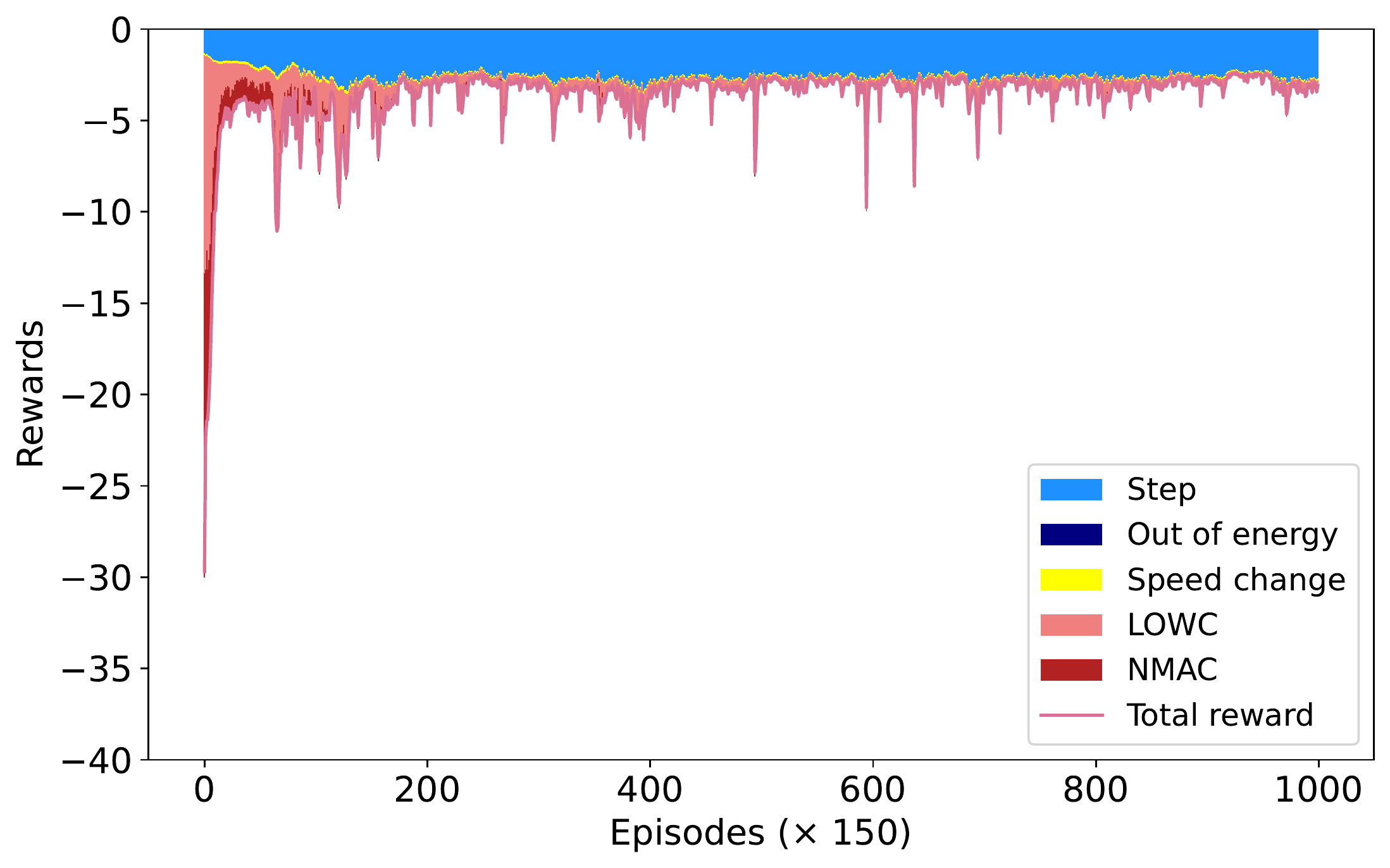}%
\label{fig_second_case}}
\hfil
\subfloat[]{\includegraphics[width=\columnwidth]{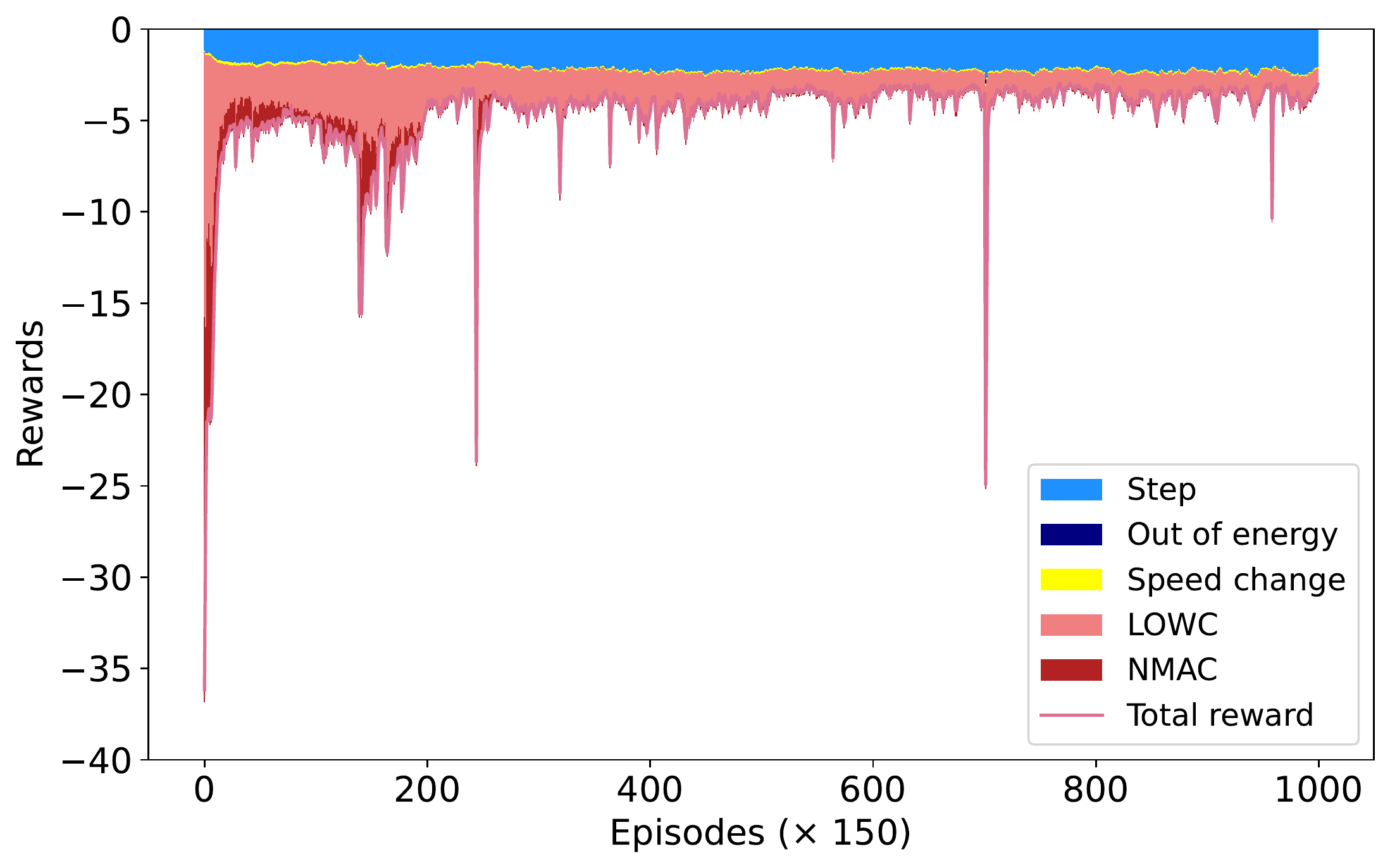}%
\label{fig_third_case}}
\hfil
\subfloat[]{\includegraphics[width=\columnwidth]{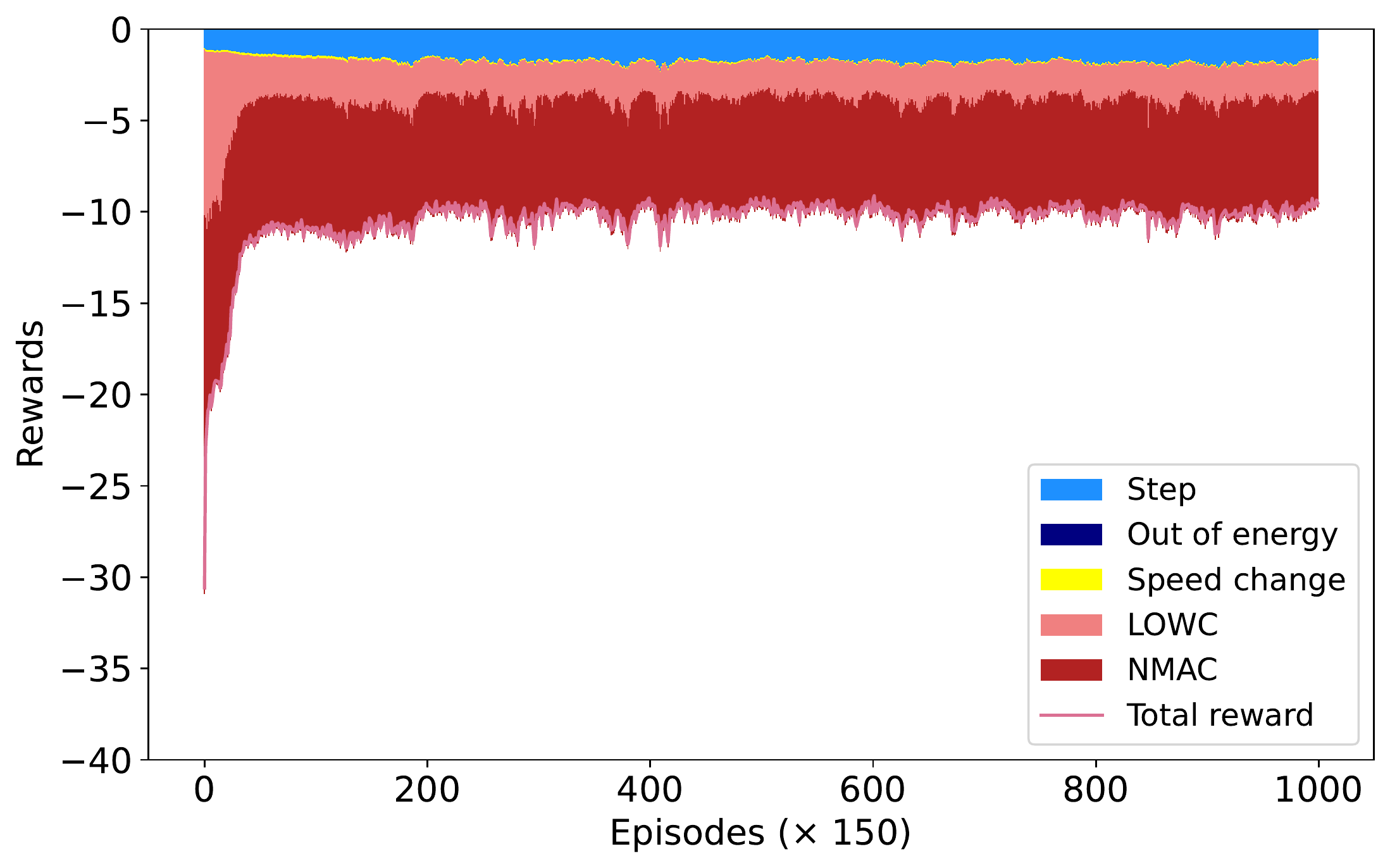}%
\label{fig_forth_case}}
\caption{MARL learning curve for different capacities. (a) Capacity=6 operations per 200s window. (b) Capacity=8 operations per 200s window. (c) Capacity=10 operations per 200s window. (d) Without DCB.}
\label{fig_sim}
\end{figure*}

Figure \ref{fig_sim} depicts the learning curve on capacities of 6, 8, 10, and 30 operations per 200 seconds window, the latter of which corresponds to the case without DCB. 
The figure indicates that as the capacity increases, the MARL model faces greater difficulty in reaching the optimal policy. 
For instance, for a capacity of 6 operations per 200 seconds window, the model converges after 30,000 episodes, while for a capacity of 8 operations per 200s window, it continues searching for up to 120,000 episodes. 
Furthermore, the figure clearly illustrates the different components of the reward function described in Section \ref{reward function}. 
In Figure \ref{fig_first_case}, LoWC and NMAC events are infrequent, and the only cost incurred is the step penalty, which is introduced from the actual flying time and is unavoidable. 
In contrast, in Figure \ref{fig_second_case} and Figure \ref{fig_third_case}, the occurrence of NMAC is rare, while LoWC is more significant. 
Additionally, the speed change penalty is higher than in Figure \ref{fig_first_case} since the agent requires more maneuvers to avoid collisions. 
Figure \ref{fig_forth_case} shows how MARL attempts to mitigate conflicts with no preconditioning by DCB. 
The primary component is the NMAC penalty, which implies a failure policy.

After careful consideration, we selected the best model trained with a capacity of 10 operations per 200s window for the subsequent experiments. 
This is because we want a model that will seek to prevent NMACs and this is the highest capacity that results in very few NMACs. In this paper, we do not seek to minimize LoWC events. 
It is noted that a MARL model trained on a highly constrained scenario generally performs well on a scenario that is not highly constrained, but the reverse may not hold.

\subsubsection{\bf Performance with Different Capacities}

After showing the feasibility of MARL, the next challenge is to determine the maximum capacity that each tactical deconfliction method can support, while meeting a Target Level of Safety (TLS). 
To address this issue, we employed Monte Carlo simulations and evaluated system performance across a range of capacities from 1 to 11 operations per 200s window. 
Each capacity was applied for 30 simulation runs and the average value of the estimated MAC was recorded in each case. 
In order to observe the efficacy of DCB on different capacities, the original traffic demand was set up at a high level, where the average demand interval is 30 seconds on each route. 
To select the appropriate capacity, we compared the average estimated MAC  against a TLS of 0.94 MAC per 100,000 flight hours, in accordance with the United States Department of Transportation's proposed TLS for General Aviation aircraft in 2023 \cite{USDOT2022}. 

Table \ref{capacity table} displays the average estimated MACs for different capacities. 
As the capacity increases, the estimated MACs also increase for all three tactical methods, indicating that DCB can function effectively to precondition for tactical deconfliction. 
The table also reveals that, at any capacity level, the performance of the MARL model is superior to that of the rule-based approach. 
Based on the predefined TLS, we selected a capacity of 4 operations per 200s window for the system with the rule-based tactical method and a capacity of 7 operations per 200s window for the system with the MARL tactical method. This indicates that the MARL method is able to meet the TLS at a higher demand than the rule-based method.    
Furthermore, if the system lacks any tactical deconfliction method, only a capacity of 1 operation per 200s window is viable. 
This is effectively strategic deconfliction since only 1 operation is released into each time window.

\begin{table}[ht!]
\caption{estimated macs on different capacities}
\label{capacity table}
\centering
\begin{tabular}{|cc|ccc|}
\hline
\multicolumn{2}{|c|}{}                                                                                          & \multicolumn{3}{c|}{\textbf{Estimated MACs per 100,000 flight hours}}                                                           \\ \hline
\multicolumn{2}{|c|}{\textbf{Traffic demand}}                                                                  & \multicolumn{3}{c|}{\textbf{High}}                                                                              \\ \hline
\multicolumn{2}{|c|}{\textbf{\begin{tabular}[c]{@{}c@{}}Tactical method\end{tabular}}}                    & \multicolumn{1}{c|}{\textbf{$\quad$None$\quad$}}         & \multicolumn{1}{c|}{\textbf{Rule-based}}   & \textbf{MARL}         \\ \hline
\multicolumn{1}{|c|}{\multirow{11}{*}{\textbf{\begin{tabular}[c]{@{}c@{}}Capacity\\ of DCB\end{tabular}}}} & 1  & \multicolumn{1}{c|}{\textbf{0.00}}                 & \multicolumn{1}{c|}{0.00}                 & 0.00                 \\ \cline{2-5} 
\multicolumn{1}{|c|}{}                                                                                     & 2  & \multicolumn{1}{c|}{41.55} & \multicolumn{1}{c|}{0.12} & 0.00                 \\ \cline{2-5} 
\multicolumn{1}{|c|}{}                                                                                     & 3  & \multicolumn{1}{c|}{56.89} & \multicolumn{1}{c|}{0.68} & 0.00                 \\ \cline{2-5} 
\multicolumn{1}{|c|}{}                                                                                     & 4  & \multicolumn{1}{c|}{75.51} & \multicolumn{1}{c|}{\textbf{0.74}}  & 0.85 \\ \cline{2-5} 
\multicolumn{1}{|c|}{}                                                                                     & 5  & \multicolumn{1}{c|}{84.32} & \multicolumn{1}{c|}{10.30} & 0.90 \\ \cline{2-5} 
\multicolumn{1}{|c|}{}                                                                                     & 6  & \multicolumn{1}{c|}{61.52} & \multicolumn{1}{c|}{43.06} & 0.71 \\ \cline{2-5} 
\multicolumn{1}{|c|}{}                                                                                     & 7  & \multicolumn{1}{c|}{122.14} & \multicolumn{1}{c|}{115.34} &  \textbf{0.66} \\ \cline{2-5} 
\multicolumn{1}{|c|}{}                                                                                     & 8  & \multicolumn{1}{c|}{112.65} & \multicolumn{1}{c|}{242.24} & 3.70 \\ \cline{2-5} 
\multicolumn{1}{|c|}{}                                                                                     & 9  & \multicolumn{1}{c|}{145.36} & \multicolumn{1}{c|}{370.52} & 23.88 \\ \cline{2-5} 
\multicolumn{1}{|c|}{}                                                                                     & 10 & \multicolumn{1}{c|}{163.31} & \multicolumn{1}{c|}{623.08} & 23.49 \\ \cline{2-5} 
\multicolumn{1}{|c|}{}                                                                                     & 11 & \multicolumn{1}{c|}{155.52} & \multicolumn{1}{c|}{673.24} & 32.39 \\ \hline
\end{tabular}
\end{table}

\subsubsection{\bf Model Comparison}
In experiments 1 and 2 we successfully trained an effective MARL model for tactical deconfliction and established the maximum capacities of various tactical methods for strategic conflict management. 
In the experiment described here, we integrated these two components and conducted a comprehensive analysis, comparing different algorithm combinations using the six metrics outlined in section \ref{formulation}. 
To evaluate the impact of different traffic demand levels, we tested each method under high, medium, and low traffic demand levels, corresponding to average departure intervals of 30, 60, and 120 seconds on each route, respectively. 
To ensure accuracy and eliminate the effects of randomness, we ran each experiment setting 30 times and reported the average values for each metric.

The final results are presented in Table \ref{final result}. 
They lead us to draw several important conclusions.
\begin{itemize}
    \item DCB is essential for safe separation. By incorporating a suitable maximum capacity for DCB, we were able to mitigate conflicts and maintain estimated MACs under the TLS. 
    The first three rows in Table \ref{final result} do not apply DCB. 
    The first represents no tactical deconfliction, the second the rule-based tactical deconfliction method, and the third the MARL tactical deconfliction method, all applied without preconditioning by DCB to reduce the demand on the tactical systems to levels that would allow them to meet the TLS. 
    Hence we do not expect the estimated MAC per 100,000 flight hours to meet the TLS in these cases.
    The last three rows correspond to the same tactical deconfliction methods, but with the DCB applied to precondition the traffic demand to a level that will allow the tactical deconfliction method to meet the TLS. 
    It is evident that DCB plays a crucial role in eliminating conflicts and ensuring safety.
    \item DCB can help save energy by reducing fuel consumption and emissions. 
    When traffic demand is high, DCB can lower the number of alerts and shorten flying time, which improves the efficiency metrics. 
    However, to implement DCB, aircraft are delayed on the ground, with the length of the delay depending on the traffic demand and maximum capacity applied. 
    It's worth noting that ground delay is not unique to DCB and exists in all three non-DCB methods as well. 
    This is because the basic departure separation method used for tactical deconfliction in all cases also causes some small ground delays. 
    \item Advanced tactical deconfliction methods, such as MARL, can increase system capacity and increase efficiency accordingly. 
    MARL combined with DCB has similar safety metrics to the rule-based method with DCB and DCB with no tactical deconfliction, and all of these methods could guarantee safe separation. 
    However, as the maximum capacity of each resource decreases, ground delay significantly increases. 
    Thus, MARL is the most efficient method simulated because it allows for a higher airspace capacity, which ultimately leads to a decrease in ground delay.
    \item The performance of the rule-based tactical deconfliction method without DCB is worse than the no-intervention case. 
    When the traffic density is too high, the risk ratio can be greater than 1, indicating that the rule-based method can lead to a higher risk of collisions than if no intervention is made at all (i.e., induce airspace risk). 
   The rationale behind this assertion is based on the potential for aircraft to experience blockages en route in the absence of DCB regulation. 
   In scenarios where DCB is not implemented, aircraft may reach their minimum speed, leaving them with limited options to avoid collisions. 
   While it is possible to execute other rule-based tactical maneuvers to prevent blockages, our paper does not model them for the sake of simplicity.
    This observation highlights the necessity of using DCB in such scenarios, which can help reduce the risk of collisions and improve overall efficiency.
\end{itemize}

\begin{table*}[ht]
\caption{Numerical Results}
\label{final result}
\centering
\begin{tabular}{ll|lll|lll|lll}
\toprule[2pt]
\multicolumn{2}{c|}{\textbf{Traffic demand}}                                           & \textbf{High}      & \textbf{Medium}      & \textbf{Low}      & \textbf{High}      & \textbf{Medium}     & \textbf{Low}     & \textbf{High}      & \textbf{Medium}     & \textbf{Low}     \\ \midrule[1pt]
\multicolumn{2}{c|}{\textbf{Safety metrics}}                                            & \multicolumn{3}{c|}{\textbf{LoWCs/ flight hr}}                    & \multicolumn{3}{c|}{\textbf{Estimated MACs/ 100,000 flight hrs}}           & \multicolumn{3}{c}{\textbf{Risk ratio}}                     \\ \midrule[1pt]
\multicolumn{1}{l|}{\multirow{6}{*}{\textbf{Algorithm}}} & \textbf{No Intervention}     & 467.0               & 313.0                  & 162.4               &205.53             &137.62               &76.19            & -                  & -                   & -                \\
\multicolumn{1}{l|}{}                                    & \textbf{Rule-based}          & 1263.3              & 908.5                & 376.7              &908.25              &559.46              &193.73          & 4.4195             & 4.0654            &2.5423           \\
\multicolumn{1}{l|}{}                                    & \textbf{MARL}                & 792.4               & 232.5                  & 127.8               &195.51              &17.53               &30.85           & 0.9513              & 0.1274               &0.4049            \\
\multicolumn{1}{l|}{}                                    & \textbf{DCB, C=1}            & 0.0                  & 0.0                    & 0.00                 &0.00              & 0.00               & 0.00            & 0.0000                  & 0.0000                   & 0.0000                \\
\multicolumn{1}{l|}{}                                    & \textbf{Rule-based+DCB, C=4} & 25.9                 & 34.8                   & 30.8                &0.74            &0.77               &0.92          &0.0036              & 0.0056               &0.0120            \\
\multicolumn{1}{l|}{}                                    & \textbf{MARL+DCB, C=7}       & 45.6                & 62.8                  & 49.6                &0.66            &0.65            &0.51           &0.0032              & 0.0047               &0.0068           \\ \midrule[1pt]
\multicolumn{2}{c|}{\textbf{Efficiency metrics}}                                        & \multicolumn{3}{c|}{\textbf{Number of alerts}} & \multicolumn{3}{c|}{\textbf{Airborne delay (seconds)}} & \multicolumn{3}{c}{\textbf{Ground delay (seconds)}} \\ \midrule[1pt]
\multicolumn{1}{l|}{\multirow{6}{*}{\textbf{Algorithm}}} & \textbf{No Intervention}     & -                  & -                    & -                 & 0.0              & 0.0               & 0.0           & 28.7                  & 9.1                   & 3.5                \\
\multicolumn{1}{l|}{}                                    & \textbf{Rule-based}          & 73.1             & 66.0                & 50.2             & 260.3              & 191.3              & 93.6            & 28.7                  & 9.1                   & 3.5                \\
\multicolumn{1}{l|}{}                                    & \textbf{MARL}                & 25.9               & 22.8                 & 15.5              & 71.4              & 78.3               &26.4            & 28.7                  & 9.1                   & 3.5                \\
\multicolumn{1}{l|}{}                                    & \textbf{DCB, C=1}            & -                  & -                    & -                 & 0.0              & 0.0               & 0.0            & 2566.1             & 2580.9              & 2444.4           \\
\multicolumn{1}{l|}{}                                    & \textbf{Rule-based+DCB, C=4} & 22.6               & 32.1                 & 25.6              & 15.0              & 23.7               & 18.7            & 505.2              & 406.2               & 293.0            \\
\multicolumn{1}{l|}{}                                    & \textbf{MARL+DCB, C=7}       & 18.1               & 19.5                 & 16.5              & 30.8              & 32.9               & 22.2            & 158.4              & 74.4               & 19.9        \\     
\bottomrule[2pt]
\end{tabular}
\end{table*}

\subsubsection{\bf Speed Curve Analysis}
\begin{figure*}[ht]
  \centering
  \includegraphics[width=0.9\textwidth]{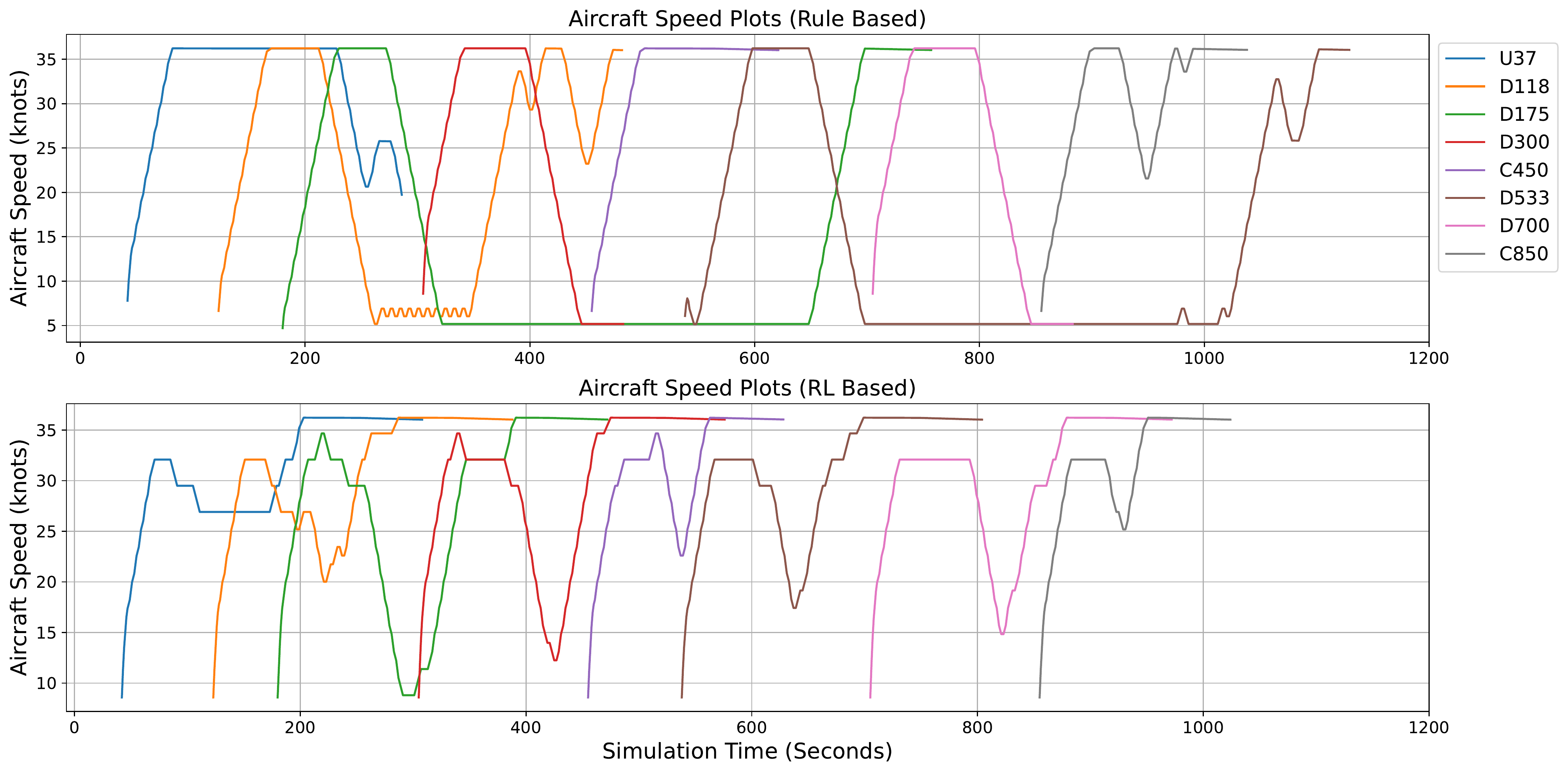}
  \caption{The comparison of speed curves of aircraft with the rule-based tactical method and the MARL methods. The y-axis in each plot represents the aircraft's actual speed in knots, while the x-axis is the simulation time in seconds. Each line represents an aircraft.}
  \label{fig:speed}
\end{figure*}

Given the differences observed between the MARL and rule-based methods for tactical deconfliction in the previous experiments, we sought to investigate the factors contributing to these differences. 
To do so, we recorded and plotted the speed curves of the simulated aircraft, as shown in Figure \ref{fig:speed}. To facilitate readability, we selected eight aircraft uniformly from the total of 30 aircraft simulated.

We observed that the rule-based method for tactical deconfliction resulted in aircraft changing speed dramatically from maximum to minimum, often with rapid acceleration and deceleration. 
In contrast, the MARL tactical deconfliction method provides speed advisories considering a longer-term view. 
For instance, for aircraft D533 (the brown curve in Figure \ref{fig:speed}), the MARL method advised holding at a relatively lower speed range for a period, helping the aircraft avoid slowing down to the minimum speed recommended by the rule-based method. 
This adjustment allowed the aircraft to arrive earlier than the rule-based method suggested. 
We also observed speed oscillations in the rule-based separation method, as illustrated by aircraft D118 (the orange curve in Figure \ref{fig:speed}). 
This occurred because the aircraft was in a situation where the distance to the leading aircraft was exactly on the boundary of the threshold for speed-up and slow-down.

In summary, the MARL tactical deconfliction method provides more optimal speed advisories compared to the rule-based method, allowing aircraft to arrive earlier and avoid rapid acceleration and deceleration, which may lead to more efficient and stable flight operations.

\section{Conclusion}\label{conclusion}
Our approach demonstrated promising results in reducing the number of conflicts and improving the efficiency of UAM operations at scale. 
The integrated conflict management framework, which combines strategic conflict management and tactical deconfliction methods, offers a comprehensive solution to address some of the challenges in high-density UAM operations. 
Our research showed that the optimization-based multiple resource demand capacity balancing algorithm plays a crucial role in preconditioning for tactical deconfliction. 
The successful implementation of game theory also improved the performance of the tactical deconfliction model, saving computational resources and making it possible to apply the system in the real world. 
In addition, the Monte-Carlo simulation we used to study the interactions between the strategic and tactical safety assurance methods provided valuable insights that can contribute to the development of more effective and efficient UAM systems in the future.

 One of the next steps in this research is to thoroughly investigate and understand the interplay between strategic and tactical conflict management methods. 
 Currently, strategic conflict management computes the optimal departure time based on a deterministic estimated flying time based on known operations. However, tactical deconfliction within the system may introduce speed changes that can affect the estimated time of arrival (ETA) at resources. 
 As airspace networks become more complex, these time differences can accumulate and result in reduced effectiveness of the preconditioning by strategic conflict management systems. 
 Therefore, formulating the ETA stochastically by considering the method of tactical deconfliction could increase the system's robustness in complex networks.

\section*{Acknowledgements}
Authors Brittain and Wei are partially supported by the NASA Grant 80NSSC21M0087 under the NASA System-Wide Safety (SWS) program.

\bibliographystyle{IEEEtran}

\bibliography{sample}

\newpage

\begin{IEEEbiography}[{\includegraphics[width=1in,height=1.25in,clip,keepaspectratio]{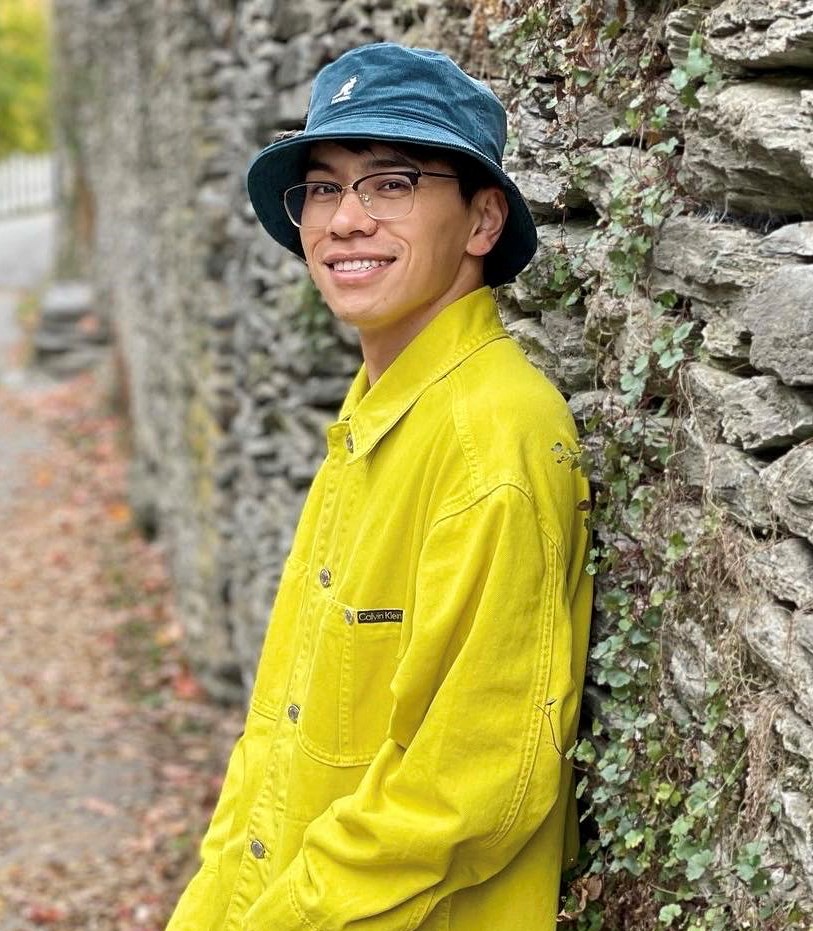}}]{Shulu Chen}
received a bachelor's degree from the School of Automation Science and Electrical Engineering at Beihang University and received Master's degree from the Department of Industrial Engineering at the University of Illinois at Urbana Champaign. He is currently pursuing the Ph.D. degree with the Department of Electrical and Computer Engineering at George Washington University. He is also working as 
a Research Assistant with the Intelligent Aerospace Systems Laboratory (IASL), under the supervision of Prof. Peng Wei. His research interests include deep reinforcement learning, optimization, and game theory, with applications in air traffic management and airline revenue management.  \end{IEEEbiography}

\begin{IEEEbiography}[{\includegraphics[width=1in,height=1.25in,clip,keepaspectratio]{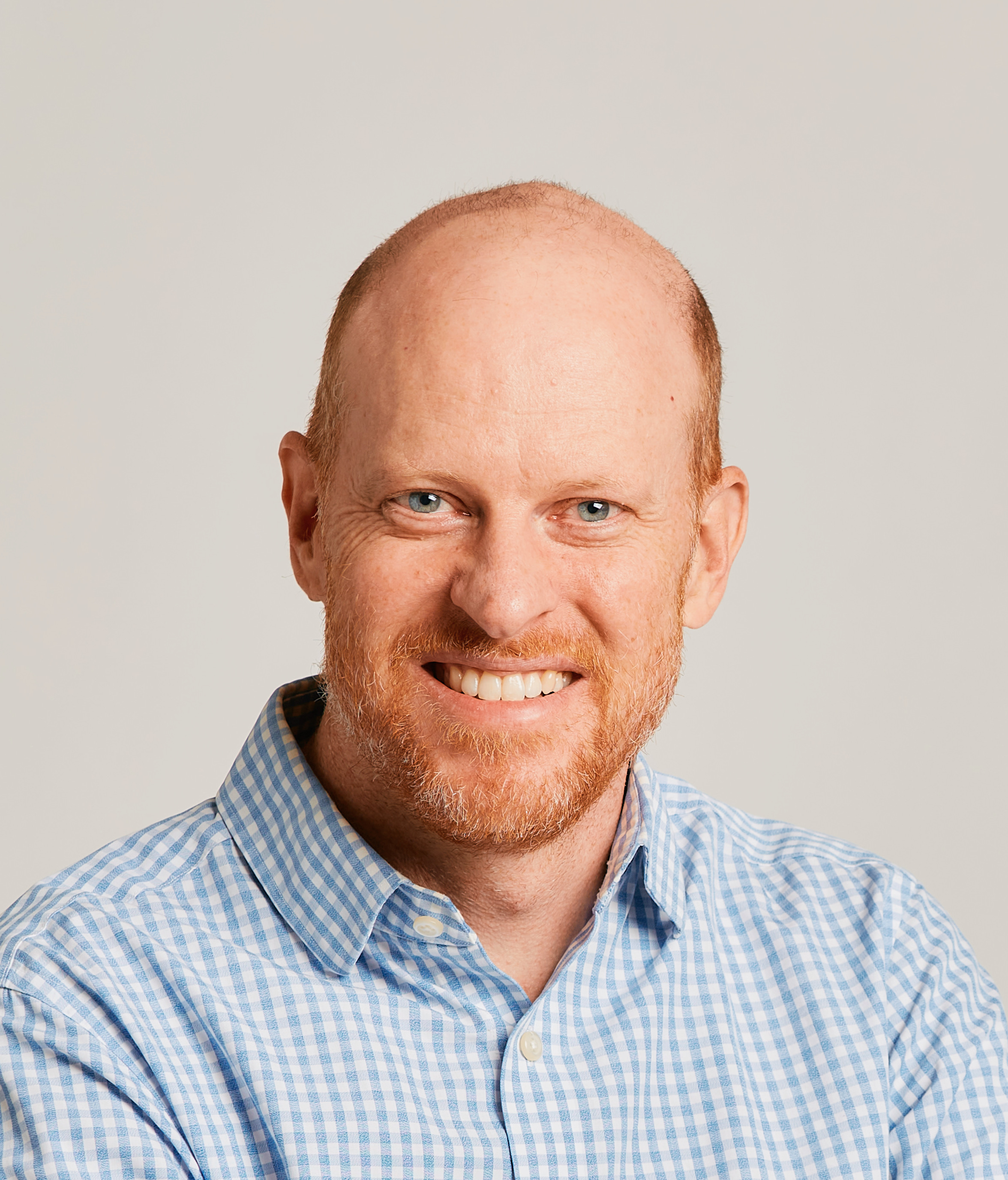}}]{Antony D. Evans} is the Director of System Design for Airbus UTM at Acubed, the Airbus innovation center in Silicon Valley, California. Tony has 17 years of research experience in air transportation, and has published widely on air traffic management, aviation and the environment, unmanned traffic management and urban air mobility. He has two Masters degrees from MIT and a PhD from the University of Cambridge. 
\end{IEEEbiography}

\begin{IEEEbiography}[{\includegraphics[width=1in,height=1.25in,clip,keepaspectratio]{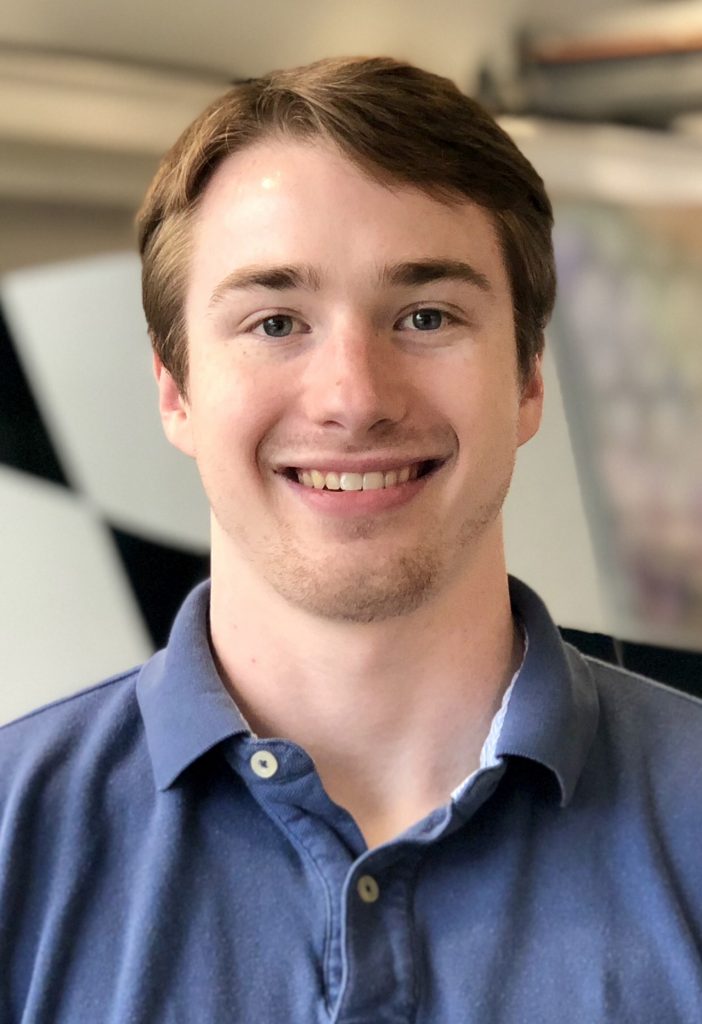}}]{Marc Brittain}
 is a member of the Technical Staff at MIT Lincoln Laboratory in the Air Traffic Control mission area. His research interests include decision making under uncertainty, safe artificial intelligence, and reinforcement learning in air transportation. At Lincoln Laboratory, his works includes the development of the AI Testbed for Advanced Air Mobility (AAM-Gym) and the evaluation of AI algorithms for separation assurance in AAM corridors. Marc serves as a member on the AIAA Air Transportation Technical Committee. He holds a Ph.D. in Aerospace Engineering from Iowa State University.
\end{IEEEbiography}

\begin{IEEEbiography}[{\includegraphics[width=1in,height=1.25in,clip,keepaspectratio]{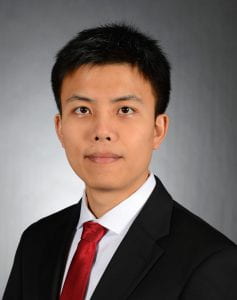}}]{Peng Wei}
 (Member, IEEE) received the Ph.D. degree in aerospace engineering from Purdue University, in 2013. He is currently an Assistant Professor
with the Department of Mechanical and Aerospace Engineering, George Washington University, with courtesy appointments at the Electrical and Computer Engineering Department and the Computer
Science Department. He is also leading the Intelligent Aerospace Systems Laboratory (IASL). By contributing to the intersection of control, optimization, machine learning, and artificial intelligence, he develops autonomy and decision support tools for aeronautics, aviation, and aerial robotics. His current research interests include safety, efficiency and scalability of decision making systems in complex, uncertain, and dynamic environments. His recent applications include Air Traffic Control/Management (ATC/M), Airline Operations, UAS Traffic Management (UTM), eVTOL Urban Air Mobility (UAM), and Autonomous Drone Racing (ADR). He is an Associate Editor of the AIAA Journal of Aerospace Information Systems.
\end{IEEEbiography}

\vfill

\end{document}